%% file: main.tex
\documentclass[letterpaper]{article} 
\usepackage[preprint]{aaai2027}  

\usepackage[hyphens]{url}  
\usepackage{graphicx} 
\usepackage{natbib}  
\usepackage{caption} 
\usepackage{algorithm}
\usepackage{algorithmic}
\usepackage{amsmath}
\usepackage{amssymb}
\usepackage{multirow}

\usepackage{newfloat}
\usepackage{listings}
\DeclareCaptionStyle{ruled}{labelfont=normalfont,labelsep=colon,strut=off} 
\floatstyle{ruled}
\newfloat{listing}{tb}{lst}{}
\floatname{listing}{Listing}

\usepackage{booktabs}

\newcommand{\vizanchorappendixdir}{VizAnchor_ArXiv_Supplement_20260825}
\input{\vizanchorappendixdir/arxiv_appendix_preamble.tex}

\title{VizAnchor: Decoding Manipulation Intent \\from Tampering Visualizations via Dual-Anchor Reasoning}
\author{
Xiaotian Zhang\equalcontrib\textsuperscript{\rm 1},
Huayuan Ye\equalcontrib\textsuperscript{\rm 2},
Haiyang Zhang\textsuperscript{\rm 1},\\
Chenhui Li\textsuperscript{\rm 3} ,
Changbo Wang\textsuperscript{\rm 1},
Sicheng Song\textsuperscript{\rm 1}\corresponding
}

\affiliations{
\textsuperscript{\rm 1}School of Data Science and Engineering, East China Normal University, Shanghai, China\\
\textsuperscript{\rm 2}Division of Emerging Interdisciplinary Areas, The Hong Kong University of Science and Technology, Hong Kong, China\\
\textsuperscript{\rm 3}School of Computer Science and Technology, East China Normal University, Shanghai, China\\
10225501407@stu.ecnu.edu.cn, hyeaq@connect.ust.hk,
51295903118@stu.ecnu.edu.cn,\\
chli@cs.ecnu.edu.cn, \{cbwang, scsong\}@dase.ecnu.edu.cn
}

\begin{document}

\maketitle

\begin{abstract}
Data visualizations are widely used for communicating information, but they are also vulnerable to intentional manipulations that induce misleading interpretations. Existing methods focus on locating tampered regions or recovering hidden information, without explaining how the visualization has been manipulated or why the resulting changes may mislead viewers. We propose \textbf{VizAnchor}, a framework for visualization manipulation understanding through dual-anchor evidence construction and VLM-based reasoning. In the first stage, VizAnchor constructs a semantic anchor to recover authentic chart information and a spatial anchor to localize tampered regions. In the second stage, three specialized agents decode the manipulation. The misleader grounding agent analyzes a four-panel visual prompt to predict the misleader information. The chart narrative reconstruction agent takes the original and tampered charts as inputs and reconstructs their respective visual narratives. Finally, the intent inferring agent integrates the visual evidence and misleader information to infer the misleading intent. We further construct a dataset for tampering localization and a dataset for misleading intent inferring. Evaluation shows that VizAnchor accurately localizes manipulations and produces faithful explanations of their manipulation, misleaders, and misleading intents.
\end{abstract}

\section{Introduction}

\begin{figure}[t]
    \centering
    \includegraphics[
        width=\columnwidth,
        keepaspectratio
    ]{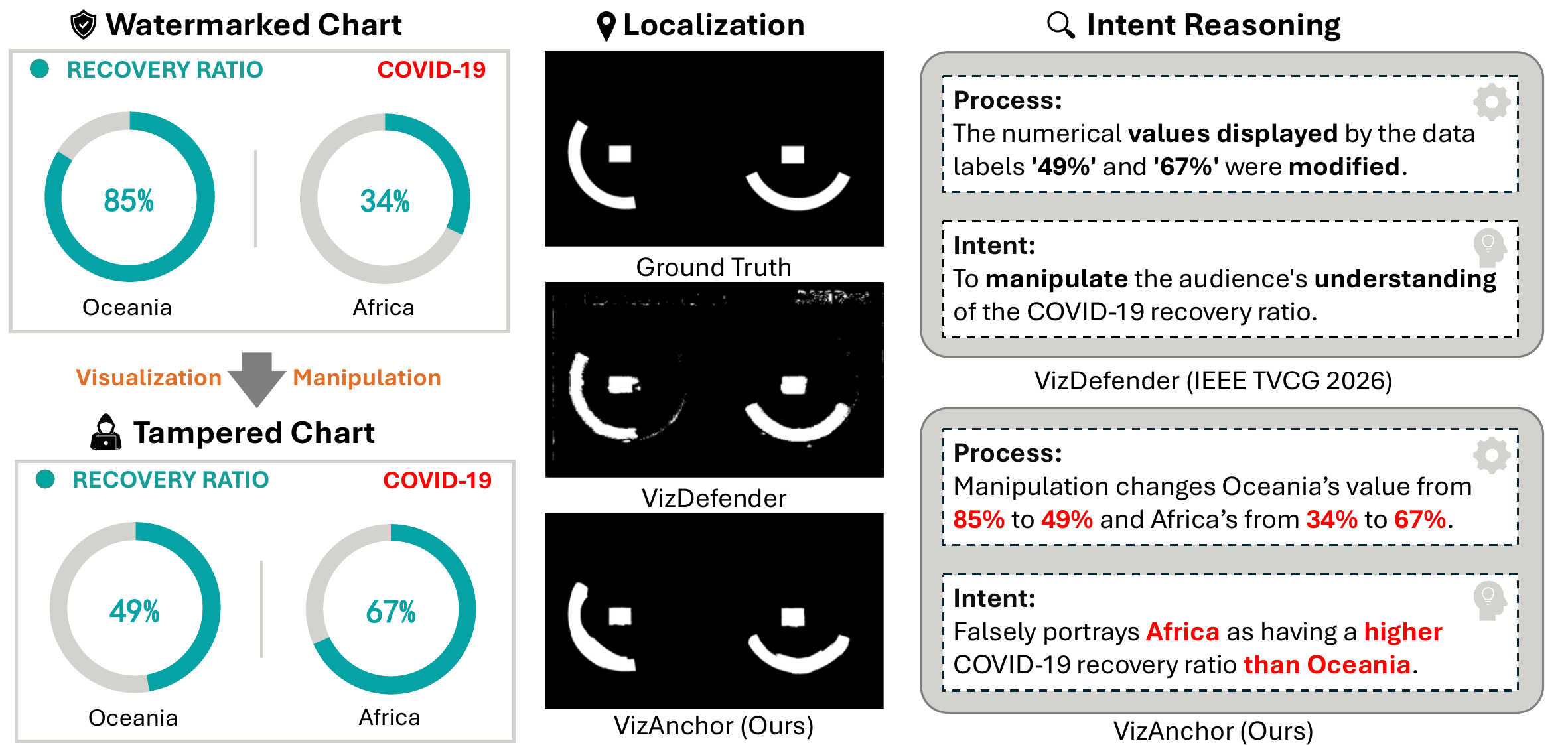}
    \caption{
    An example of manipulation intent decoding with VizAnchor.
    Swapping the category labels of Africa and Oceania reverses the COVID-19
    recovery comparison. Guided by semantic and spatial anchors, the multi-agent
    decoder grounds the manipulation, reconstructs both chart narratives, and
    infers its misleading intent.
    }
    \label{fig:example}
\end{figure}

Visualizations are widely used to communicate quantitative evidence in scientific articles, financial reports, public dashboards, and news media. Their persuasive power, however, also makes them vulnerable to subtle manipulation. Small edits to data marks, axes, legends, labels, colormaps, or annotations can alter the conclusions drawn by viewers while preserving a visually plausible appearance. Unlike generic image tampering, visualization manipulation is inherently semantic, as its impact depends on how visual edits change data encodings, comparisons, and communicated messages.

Existing trustworthy-visualization methods mainly focus on provenance verification and data recovery. Methods such as VisCode, InvVis, and VisGuard embed metadata, source data, or recoverable information into visualization images, enabling users to trace a chart back to its original data or generation context~\cite{zhang2021viscode,ye2024invvis,ye2026visguard}. However, they are not designed to precisely localize malicious edits or explain their semantic consequences. VizDefender further combines tamper localization with multimodal reasoning~\cite{song2026vizdefender}, but its analysis relies on the tampered chart and detected suspicious regions. Without an authentic visual reference, it remains difficult to determine how the manipulated chart deviates from its original message and why the resulting interpretation is misleading.

To address these limitations, as shown in Figure~\ref{fig:example}, we introduce \textbf{VizAnchor}, a two-stage framework for fine-grained visualization manipulation understanding. In Stage 1, VizAnchor embeds recoverable provenance metadata and a crop-synchronization cue into an original chart, producing a watermarked chart. Given a potentially cropped and manipulated chart, VizAnchor first decodes its position map and restores it to the canonical canvas, producing aligned tampered chart. Metadata is then recovered from the aligned chart and used to retrieve the original chart and untampered watermarked reference. The recovered original chart serves as the Semantic Anchor, while the localization module compares with aligned tampered chart to construct the Spatial Anchor.

In Stage 2, the dual anchors are organized into a four-panel visual prompt containing Semantic Anchor, tampered chart, a Tamper Map, and a Localized Comparison. The Misleader Grounding Agent predicts the tamper type, tampered components, and manipulation process from this prompt. The Chart Narrative Reconstruction Agent separately analyzes
Semantic Anchor and tampered chart to reconstruct their respective
chart narratives. Finally, the Intent Inferring Agent integrates the visual prompt and outputs of the preceding agents to infer the underlying misleading intent.

Our contributions are summarized as follows:

\begin{itemize}
    \item We propose VizAnchor, a framework for visualization manipulation analysis that integrates provenance recovery, spatial alignment, tampering localization, and misleading-intent decoding.

    \item To address the lack of authentic references and precise spatial evidence in existing visualization manipulation analysis, we introduce a dual-anchor evidence construction mechanism that combines a provenance-verified Semantic Anchor with a pixel-level Spatial Anchor.

    \item We develop a multi-agent VLM reasoning framework that progressively grounds manipulation attributes, reconstructs original and tampered chart narratives, and infers misleading intent from structured visual evidence.

    \item We construct two datasets for localization training and manipulation-understanding evaluation. Extensive experiments demonstrate reliable watermark recovery, accurate crop and local-edit localization, and effective reasoning about manipulation processes and misleading intents.
\end{itemize}

\section{Related Work}

\paragraph{Misleading Visualization.}
Prior work has examined how distorted scales, encodings, visual
proportions, and rhetorical designs can bias chart interpretation
~\cite{pandey2015deceptive,lo2022misinformed,lisnic2023misleading,
lisnic2024yeah}. Recent benchmarks further evaluate whether MLLMs can
detect or explain misleading cues in standalone charts
~\cite{chen2025unmasking,mahbub2025perils}. Most of these works focus on misleaders introduced during the initial chart-creation process, which may arise from either deliberate design choices or unintentional design flaws.

We regard tampered visualizations as a more specific subset of
misleading visualizations. In this setting, an initially valid chart is intentionally modified after creation at the image level to alter its original message. Therefore, the task requires not only recognizing that the resulting chart is misleading, but also identifying what was changed relative to the authentic chart and inferring the intent behind the modification. Existing misleading-chart benchmarks, such as Misviz~\cite{tonglet2026is} and Misleading ChartQA~\cite{chen2025unmasking}, provide complementary evaluation of general misleading-chart understanding, but do not directly target post-creation tampering intent. VizAnchor addresses this gap by reasoning from authentic and tampered visualization evidence to localize the edit, explain the manipulation process, and infer its misleading intent.

\paragraph{Visualization Protection.}
Existing visualization protection methods generally follow two
directions. One line embeds chart metadata through robust watermarking to support provenance verification and source-data recovery~\cite{fu2021chartem,zhang2021viscode,ye2024invvis}. Another line embeds location-aware or semi-fragile signals to detect and localize post-creation tampering~\cite{ye2026visguard,song2026vizdefender}. However, provenance recovery and tamper localization alone do not fully explain how an edit changes the chart message or what misleading intent it serves. VizAnchor bridges these objectives by combining dual-anchor evidence with VLM-based agents, enabling provenance retrieval, tamper localization, and fine-grained manipulation-intent analysis within a unified framework.

\paragraph{Image Tampering Detection.}
Image tampering detection is commonly divided into passive and active settings. Passive tampering detectors localize edits from forensic traces such as noise residuals, boundary artifacts, and cross-channel inconsistencies~\cite{zhou2018learning,wu2019mantra,chen2021mvss,liu2022pscc, guillaro2023trufor}. These methods are powerful for generic image manipulation, but they are not tailored to visualization semantics, where small edits to axes, legends, labels, or marks may cause large interpretive shifts. Active methods instead embed signals before dissemination~\cite{zhu2018hidden,tancik2020stegastamp,zhang2019rivagan,jia2021mbrs, kou2025iwrn}, with EditGuard and OmniGuard extending this paradigm to proactive tamper localization~\cite{zhang2024editguard,zhang2025omniguard}. Unlike generic image tampering detection, VizAnchor targets visualization manipulation understanding, localization is used as spatial evidence for reasoning about misleaders and misleading effects.

\begin{figure*}[t]
    \centering
    \includegraphics[width=\textwidth]{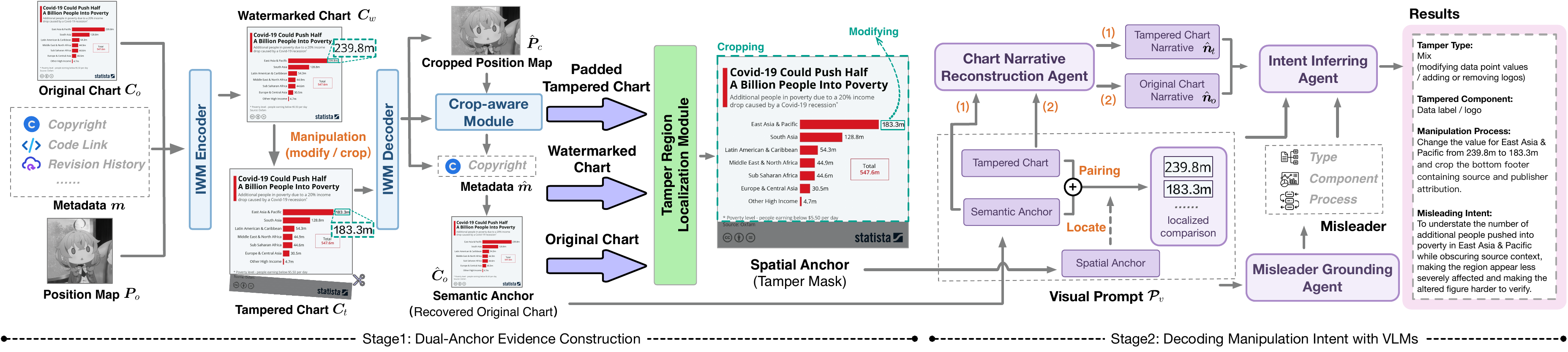}
    \caption{Overview of the proposed VizAnchor framework.}
    \label{fig:framework}
\end{figure*}

\section{Method}
\label{sec:method}


VizAnchor moves beyond tamper localization to understand how a chart is manipulated and why it becomes misleading. As shown in
Figure~\ref{fig:framework}, it consists of two stages:
Dual-Anchor Evidence Construction and Decoding Manipulation Intent with VLMs.

\subsection{Semantic Anchor Construction}
\label{subsec:semantic_anchor_construction}

The Semantic Anchor is the recovered original chart $\hat{C}_o$,
which provides the semantic reference for comparison with the aligned tampered chart $\widetilde{C}_t$. To bind this reference to the distributed image, VizAnchor performs crop-robust metadata hiding and produces the watermarked chart $C_w$.

\begin{figure}[t]
    \centering
    \includegraphics[width=\columnwidth]{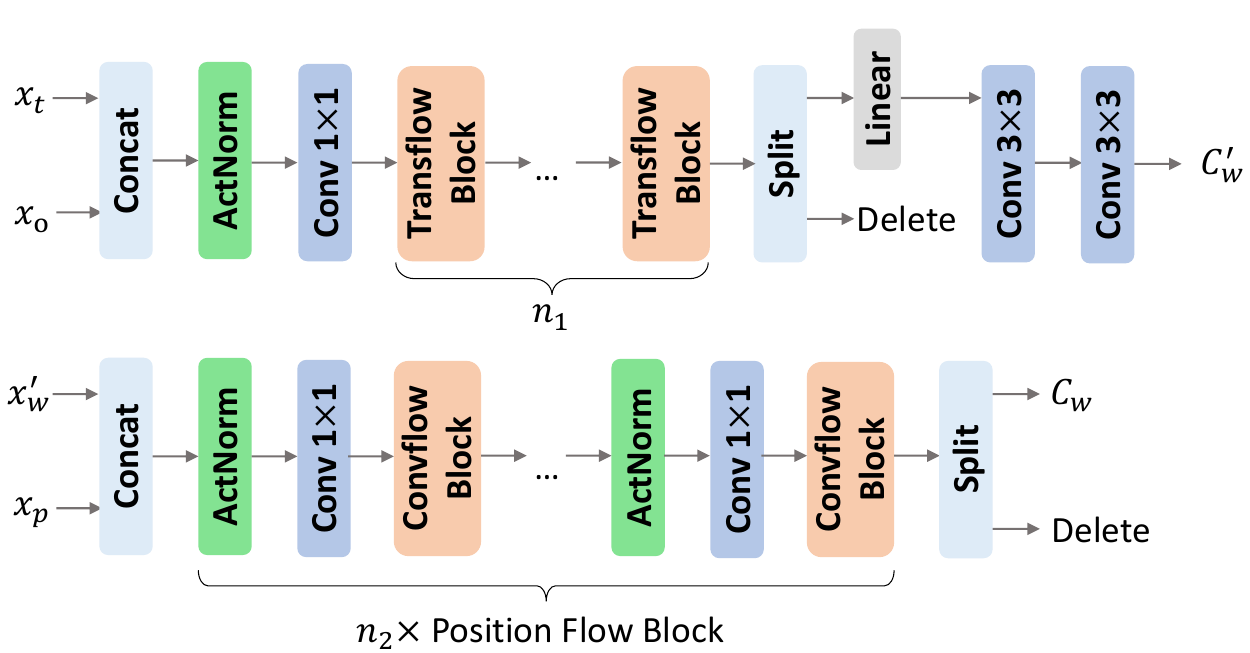}
    \caption{
    Architecture of the Invertible Watermarking Module. The IWM Encoder embeds metadata into the chart through Transflow Blocks to produce the intermediate watermarked chart $C'_w$, and then embeds the position map through Position Flow Blocks to obtain $C_w$. The IWM Decoder reverses these stages for position and metadata recovery.
    }
    \label{fig:iwm_block}
\end{figure}

\paragraph{Metadata Embedding and Recovery.}
As illustrated in Figure~\ref{fig:iwm_block}, IWM is an INN-based reversible network built upon affine coupling transformations~\cite{dinh2017density}. Following the attention flow-based message embedding design of RMSteg~\cite{ye2025robust}, IWM sequentially embeds metadata and position information into an original chart. Its forward and inverse paths are referred to as the IWM Encoder and IWM Decoder, respectively.

Given an original chart $C_o$ and a $K$-bit metadata payload $m$, the payload is reshaped into a binary map and spatially tiled to introduce redundancy. The chart and tiled metadata map are separately tokenized using ViT-based encoders~\cite{dosovitskiy2021vit}. Unlike the chart features, the metadata features are additionally processed by an invertible token shuffle, which disperses adjacent metadata tokens across different spatial positions and improves robustness against localized corruption. The two feature streams are then concatenated and mixed through ActNorm and an invertible $1\times1$ convolution before entering a sequence of Transflow Blocks. After the final block, the chart feature stream is projected back to image space to produce the intermediate watermarked chart $C'_w$.

Let $x_t^i$ and $x_o^i$ denote the shuffled tiled-metadata features and original-chart features entering the $i$-th Transflow Block, respectively. Its forward transformation is
\begin{equation}
\begin{aligned}
x_o^{i+1}
&=
x_o^i+\phi_{\mathrm{T}}(x_t^i),\\
x_t^{i+1}
&=
x_t^i\odot
\exp\!\left(\rho_{\mathrm{T}}(x_o^{i+1})\right)
+\eta_{\mathrm{T}}(x_o^{i+1}),
\end{aligned}
\label{eq:transflow_forward}
\end{equation}
where $\phi_{\mathrm{T}}$, $\rho_{\mathrm{T}}$, and
$\eta_{\mathrm{T}}$ are learnable transformations. The corresponding inverse transformation is
\begin{equation}
\begin{aligned}
\hat{x}_t^i
&=
\left[
\hat{x}_t^{i+1}
-\eta_{\mathrm{T}}(\hat{x}_o^{i+1})
\right]
\oslash
\exp\!\left(\rho_{\mathrm{T}}(\hat{x}_o^{i+1})\right),\\
\hat{x}_o^i
&=
\hat{x}_o^{i+1}
-\phi_{\mathrm{T}}(\hat{x}_t^i),
\end{aligned}
\label{eq:transflow_inverse}
\end{equation}
where $\odot$ and $\oslash$ denote element-wise multiplication and division, respectively.

Once the aligned tampered chart $\widetilde{C}_t$ is obtained by the position-recovery stage described below, it is directly fed into the metadata-decoding branch. The inverse Transflow Blocks recover the shuffled metadata features, after which the inverse token shuffle restores their original spatial ordering. The reordered features are decoded into the tiled metadata map, and the redundant predictions are aggregated to obtain the recovered payload $\hat{m}$.

\paragraph{Position Embedding and Recovery.}
The position-embedding stage embeds the full position map $P_o$ into the intermediate watermarked chart $C'_w$. Their features are concatenated and processed by a sequence of Position Flow Blocks, producing the final watermarked chart $C_w$. Let $z_w^j$ and $z_p^j$ denote the features of $C'_w$ and $P_o$ entering the $j$-th Position Flow Block, respectively.

Within each Position Flow Block, the Convflow Block follows the same forward and inverse affine-coupling formulations as the Transflow Block in Eqs.~\eqref{eq:transflow_forward} and \eqref{eq:transflow_inverse}. However, its learnable transformations, $\phi_{\mathrm{P}}$, $\rho_{\mathrm{P}}$, and
$\eta_{\mathrm{P}}$, are parameterized independently from those used for metadata embedding. This separation enables metadata and positional information to be modeled by distinct transformations.

During recovery, given a potentially cropped and manipulated chart $C_t$, a Nested U-Net first enhances its features, and the Crop-Aware Module produces the aligned tampered chart $\widetilde{C}_t$. The aligned chart is then passed to the metadata-decoding branch to recover $\hat{m}$.

\paragraph{Crop-Aware Module.}
Given a potentially cropped and manipulated chart $C_t$, the IWM first recovers a cropped position map $\hat{P}_c$. The Crop-Aware Module matches $\hat{P}_c$ against the full position map $P_o$ to estimate the crop location and scale. The estimated transformation is then used to restore $C_t$ to the canonical canvas, producing the aligned tampered chart $\widetilde{C}_t$. Meanwhile, the missing region outside the estimated crop boundary is converted into a crop mask $\hat{M}_{\mathrm{crop}}$. The aligned chart is subsequently passed back to the inverse path of IWM for metadata recovery, while $\hat{M}_{\mathrm{crop}}$ is retained as part of the spatial evidence.

\paragraph{Semantic-Anchor Retrieval.}
Before dissemination, each metadata payload $m$ is registered with its corresponding original chart $C_o$ and untampered watermarked chart $C_w$ in a trusted repository. After crop-aware alignment, the recovered metadata $\hat{m}$ is used as a lookup key to retrieve the registered pair $(\hat{C}_o,\hat{C}_w)$. The recovered original chart
$\hat{C}_o$ serves as the Semantic Anchor, providing authentic chart content for downstream VLM reasoning. The untampered watermarked chart $\hat{C}_w$ is paired with the aligned tampered chart $\widetilde{C}_t$ for local-edit localization. 

\subsection{Spatial Anchor Construction}
\label{subsec:spatial_anchor_construction}

The Spatial Anchor integrates two complementary localization cues: the crop mask $\hat{M}_{\mathrm{crop}}$ produced by the Crop-Aware Module and the local-edit mask $\hat{M}_{\mathrm{edit}}$ predicted by the Localization Module. Both masks are represented on the canonical canvas and combined by pixel-wise union:
\begin{equation}
\hat{M}
=
\hat{M}_{\mathrm{crop}}
\lor
\hat{M}_{\mathrm{edit}},
\label{eq:spatial_anchor}
\end{equation}
where $\lor$ denotes the pixel-wise logical OR operation. The resulting mask $\hat{M}$ serves as the Spatial Anchor, capturing both geometric cropping and fine-grained content modifications.

\paragraph{Crop Localization.}
As described by the Crop-Aware Module under \textit{Semantic Anchor Construction}, the recovered position map $\hat{P}_c$ is matched against the full position map $P_o$ to estimate the crop region and align the observed chart to the canonical canvas. The estimated cropped-away region is represented as the binary crop mask $\hat{M}_{\mathrm{crop}}$.

\paragraph{Local-Edit Localization.}
After crop-aware alignment and reference retrieval, the Localization Module compares the untampered watermarked chart $\hat{C}_w$ with the aligned tampered chart $\widetilde{C}_t$. Its input is a seven-channel tensor consisting of the two RGB charts and their binary difference map:
\begin{equation}
\begin{aligned}
D_{\mathrm{bin}}
&=
\mathbb{I}
\left(
\max_{c\in\{R,G,B\}}
\left|
\hat{C}_w^{(c)}-\widetilde{C}_t^{(c)}
\right|
>
\tau_{\mathrm{diff}}
\right),\\
X
&=
\operatorname{Concat}
\left(
\hat{C}_w,\widetilde{C}_t,D_{\mathrm{bin}}
\right).
\end{aligned}
\label{eq:localization_input}
\end{equation}

The Localization Module adopts a U-Net-based architecture to jointly model the paired RGB content and changed-pixel cues~\cite{ronneberger2015unet}. The local-edit mask is obtained as
\begin{equation}
\hat{M}_{\mathrm{edit}}
=
\mathbb{I}
\left(
F_{\theta}(X)>\tau_{\mathrm{mask}}
\right),
\label{eq:local_edit_mask}
\end{equation}
where $F_{\theta}(X)$ denotes the predicted local-edit probability map. The threshold values are reported in the appendix.

\subsection{Training Objectives}
The IWM and Crop-Aware Module are jointly trained on the VisGuard
dataset, while the Localization Module is trained separately on VAD-LocTrain, the localization-training subset of VizAnchor Dataset (VAD), which contains 1,500 automatically generated chart pairs with pixel-level local-edit masks. The joint objective for IWM and crop-aware recovery is
\begin{equation}
\begin{aligned}
\mathcal{L}_{\mathrm{IWM}}
={}&
\lambda_{\mathrm{steg}}\mathcal{L}_{\mathrm{steg}}
+\lambda_{\mathrm{ssim}}\mathcal{L}_{\mathrm{ssim}}
+\lambda_{\mathrm{meta}}\mathcal{L}_{\mathrm{meta}} \\
&+
\lambda_{\mathrm{lpips}}\mathcal{L}_{\mathrm{lpips}}
+\lambda_{\mathrm{pf}}\mathcal{L}_{\mathrm{pf}},
\end{aligned}
\end{equation}
where $\mathcal{L}_{\mathrm{steg}}$ is the L1 loss between the original and watermarked charts, $\mathcal{L}_{\mathrm{ssim}}=1-\mathrm{SSIM}$, $\mathcal{L}_{\mathrm{meta}}$ is the BCEWithLogits loss for metadata recovery, $\mathcal{L}_{\mathrm{lpips}}$ is the VGG-based LPIPS perceptual loss, and $\mathcal{L}_{\mathrm{pf}}$ is the L1 loss between the predicted and target position maps. The Localization Module is optimized using
\begin{equation}
\mathcal{L}_{\mathrm{edit}}
=
\mathcal{L}_{\mathrm{BCE}}
+
\mathcal{L}_{\mathrm{Dice}}.
\end{equation}
All modules are trained using a single NVIDIA RTX PRO 6000 GPU. The loss-balancing coefficients and implementation details are reported in the appendix.

\subsection{Decoding Manipulation Intent with VLMs}
\label{subsec:vlm_reasoning}

After constructing the Semantic and Spatial Anchors, VizAnchor decodes manipulation intent with VLMs through a structured multi-agent reasoning pipeline. The pipeline first converts the authentic chart $\hat{C}_o$, the tampered chart $C_t$, and the predicted tamper mask $\hat{M}$ into a four-panel visual prompt. It then uses three specialized VLM agents to progressively ground visual edits, reconstruct chart narratives, and infer the misleading intent.

\paragraph{Visual Prompt Construction.}
Given $\hat{C}_o$, $C_t$, and $\hat{M}$, we construct a four-panel visual prompt
$\mathcal{P}_v=\langle \hat{C}_o, C_t, \hat{C}_m, \hat{C}_l\rangle$.
The first two panels show the authentic and tampered charts. The third panel, denoted as the Tamper Map $\hat{C}_m$, overlays the Spatial Anchor on $\hat{C}_o$ by drawing a thin contour around the tamper region. This highlights localized evidence while preserving the semantic context of the authentic chart. The fourth panel, denoted as the Localized Comparison $\hat{C}_l$, crops and enlarges the regions from $\hat{C}_o$ and $C_t$ according to $\hat{M}$ and places them side by side. This prompt provides both global chart context and fine-grained local comparison.

\paragraph{Misleader Grounding Agent.}
MGA grounds visual differences into structured manipulation semantics. Given $\mathcal{P}_v$, it predicts the tamper type, identifies the tampered components, and generates a manipulation-process description:
\[
A_{\mathrm{MGA}}(\mathcal{P}_v)
\rightarrow
\left\{
\hat{y}_{\mathrm{type}},
\hat{\mathcal{C}}_{\mathrm{comp}},
\hat{y}_{\mathrm{proc}}
\right\}.
\]
Following VizDefender~\cite{song2026vizdefender}, we adopt its taxonomy of nine chart-specific tamper types and seven tampered-component categories. The predicted type and components specify what misleader is used and which visual element is affected, while the process description explains how the tampered chart deviates from the authentic one.

\paragraph{Chart Narrative Reconstruction Agent.}
CNRA analyzes how the chart message changes after manipulation. It receives $\hat{C}_o$ and $C_t$ and reconstructs two chart narratives:
\[
A_{\mathrm{CNRA}}
\left(
\hat{C}_o,{C}_t
\right)
\rightarrow
\left\{
\hat{n}_o,\hat{n}_t
\right\},
\]
where $\hat{n}_o$ summarizes the authentic trends, comparisons, rankings, or conclusions, and $\hat{n}_t$ summarizes the message conveyed by the tampered chart. This agent captures the semantic shift caused by the manipulation, complementing the localized evidence extracted by MGA.

\paragraph{Intent Inferring Agent.}
IIA performs the final reasoning step. It receives the visual prompt $\mathcal{P}_v$, the grounded manipulation outputs from MGA, and the reconstructed narratives from CNRA:
\[
A_{\mathrm{IIA}}
\left(
\mathcal{P}_v,
\hat{y}_{\mathrm{type}},
\hat{\mathcal{C}}_{\mathrm{comp}},
\hat{y}_{\mathrm{proc}},
\hat{n}_o,
\hat{n}_t
\right)
\rightarrow
\hat{y}_{\mathrm{intent}}.
\]
The output $\hat{y}_{\mathrm{intent}}$ describes how the manipulation may mislead viewers or alter their interpretation of the chart. By integrating localized visual evidence, structured manipulation semantics, and original--tampered narrative contrast, IIA infers evidence-supported misleading strategies rather than unobservable subjective intent.

\section{Experiments}
\label{sec:experiments}

\begin{figure*}[t]
    \centering
    \includegraphics[width=\textwidth]{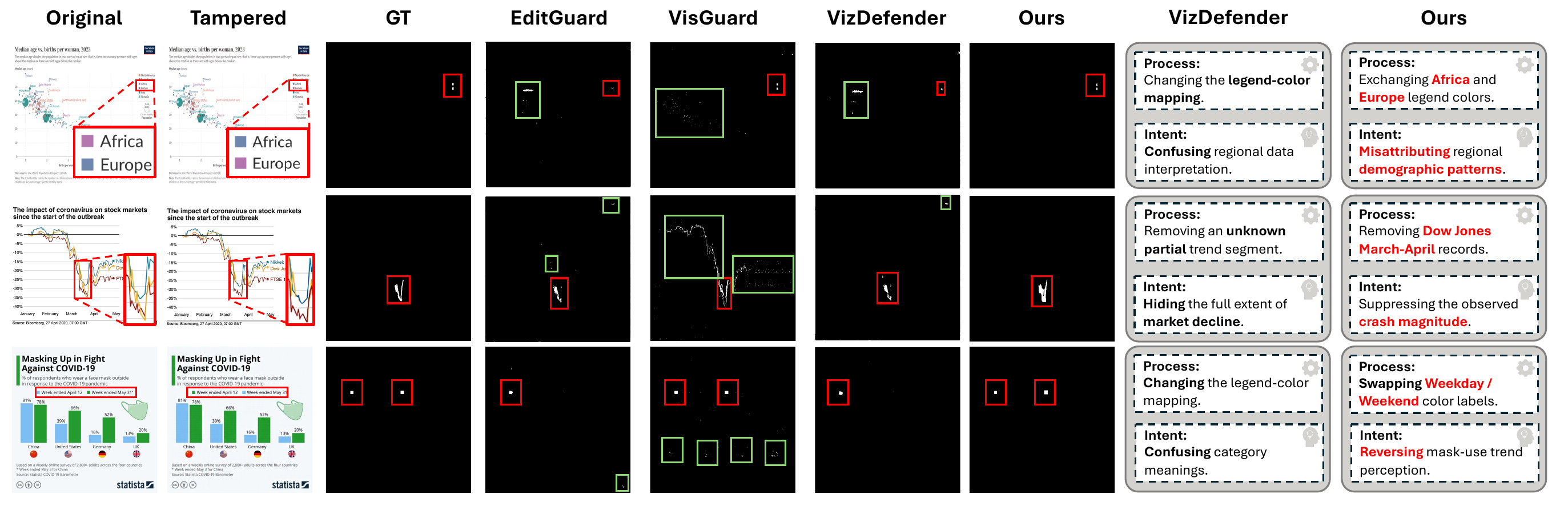}
    \caption{Qualitative comparison of tamper localization and
misleading-intent inference. Red boxes indicate correctly localized tampered regions, while green boxes highlight noise regions falsely identified as tampering.}
    \label{fig:localization_intent_comparison}
\end{figure*}

Our experiments follow the two-stage design of VizAnchor. We first evaluate the dual-anchor evidence construction, including
watermarked-chart fidelity, metadata recovery, crop localization, and tamper localization. We then evaluate VLM-based manipulation understanding, including tamper type classification, tampered component recognition, manipulation-process description, and misleading-intent inference. Finally, we conduct ablation studies to analyze the contribution of dual-anchor visual evidence and the proposed multi-agent reasoning design. Figure~\ref{fig:localization_intent_comparison} presents comparisons of tamper localization and misleading-intent inference, complementing the quantitative evaluations reported below.

\subsection{Experimental Details}
\label{subsec:exp_setup}

\paragraph{Datasets.}
We conduct experiments on VisGuard dataset (VGD)~\cite{ye2026visguard}, VizDefender dataset (VDD)~\cite{song2026vizdefender}, and our constructed datasets (VAD). Our VAD contains 1,500 automatically generated chart pairs with pixel-level masks for training the Localization Module and 120 manually created chart pairs for evaluating localization and manipulation understanding. Compared with VDD-Eval, VAD-Eval additionally contains crop-based manipulations.

\paragraph{Baselines.}
We compare VizAnchor with image-hiding, tamper-localization and manipulation-understanding methods, including HiNet, ISN, StampOne, StegaStamp, WAM, EditGuard, OmniGuard, VizDefender, and VisGuard~\cite{
jing2021hinet,
lu2021isn,
shadmand2024stampone,
tancik2020stegastamp,
sander2025watermark,
zhang2024editguard,
zhang2025omniguard,
song2026vizdefender,
ye2026visguard}.

\paragraph{Metrics.}

We evaluate watermarked chart fidelity using PSNR, SSIM~\cite{wang2004ssim}, and LPIPS~\cite{zhang2018lpips}, and assess metadata recovery reliability via bit accuracy (BitAcc). For crop and local-edit localization, we report Intersection over Union (IoU), Precision, Recall, and F1. Tamper-type classification and tampered-component recognition are evaluated using Accuracy, Exact Match Accuracy requiring the predicted component set to perfectly match the reference, and Macro-F1. For free-form manipulation-process and misleading-intent descriptions, we compute Cos-FA, the cosine similarity between text-embedding-3-large embeddings of the prediction and reference. To comprehensively assess factual correctness and consistency with visual evidence, we additionally report AI-FA, an LLM-as-a-judge metric using Gemini-3.1-Pro-Preview~\cite{liu2023geval,zheng2023judging} to score outputs from 0 to 5, which are normalized to 0 to 1. The complete evaluation prompt and scoring rubric are provided in the appendix.

\subsection{Semantic Anchor Construction}
\label{subsec:semantic_anchor_eval}

We evaluate whether the underlying IWM encoder and crop-aware decoder preserve chart appearance and recover a sufficiently reliable retrieval key under clean, tampered and cropped conditions.

\paragraph{Watermarked-Chart Fidelity.}
We evaluate watermarked-chart fidelity on VGD. As shown in Table~\ref{tab:image_quality}, VizAnchor achieves the
highest PSNR of 43.28, while its SSIM and LPIPS remain competitive with the best-performing baselines. This indicates that jointly embedding metadata and position information introduces limited visual distortion.

\paragraph{Metadata Recovery.}
We report clean-image BitAcc and robustness under two forms of corruption. For local tampering, the manipulated area occupies 15\%, 30\%, 45\%, or 60\% of the chart. For cropping, the crop ratio ranges from 10\% to 50\%. Table~\ref{tab:modified_pixel_recovery} and Table~\ref{tab:crop_metadata_recovery} show that VizAnchor consistently
achieves the highest metadata recovery accuracy under both local
modification and cropping. BitAcc remains 99.43\% when 60\% of the
pixels are modified and 99.36\% when 50\% of the chart area is cropped, demonstrating reliable Semantic Anchor retrieval under severe content loss.

\begin{table}[t]
\centering
\small
\setlength{\tabcolsep}{3.0pt}
\begin{tabular*}{\columnwidth}{@{\extracolsep{\fill}}cccc}
\toprule
Method & PSNR$\uparrow$ & SSIM$\uparrow$ & LPIPS$\downarrow$ \\
\midrule
HiNet (81)       & 29.95 & 0.8306 & 0.3692 \\
ISN (81)         & 34.38 & 0.9478 & 0.2340 \\
StampOne (81)    & 29.44 & 0.9107 & 0.1913 \\
StegaStamp (81)  & 10.32 & 0.5733 & 0.6852 \\
WAM (32)         & 31.55 & 0.9440 & 0.1513 \\
EditGuard (64)   & 33.00 & 0.8752 & 0.2416 \\
OmniGuard (100)  & \underline{41.64} & \textbf{0.9839}
                                  & \textbf{0.0443} \\
VizDefender (64) & 34.46 & 0.8840 & 0.2207 \\
VisGuard (81)    & 41.13 & 0.9734 & \underline{0.0610} \\
VizAnchor (81)   & \textbf{43.28} & \underline{0.9809} & 0.0664 \\
\bottomrule
\end{tabular*}
\caption{Image quality on VGD. The number in parentheses denotes the payload length in bits. Bold and underlined values indicate the best and second-best results, respectively.}
\label{tab:image_quality}
\end{table}

\begin{table}[t]
\centering
\small
\setlength{\tabcolsep}{2.0pt}
\begin{tabular*}{\columnwidth}{@{\extracolsep{\fill}}cccccc}
\toprule
\multirow{2}{*}{Method}
& \multicolumn{5}{c}{Modified-Pixel Ratio (\%)} \\
\cmidrule(lr){2-6}
& 0 & 15 & 30 & 45 & 60 \\
\midrule
HiNet (81)
& 98.99 & 97.87 & 94.68 & 87.29 & 74.99 \\

ISN (81)
& 99.08 & 98.20 & 96.73 & 94.49 & 91.22 \\

StampOne (81)
& 91.90 & 91.07 & 89.95 & 88.17 & 85.21 \\

StegaStamp (81)
& 98.45 & 98.07 & 97.56 & 96.61 & \underline{94.72} \\

WAM (32)
& 92.36 & 90.87 & 88.01 & 83.50 & 75.98 \\

EditGuard (64)
& 83.11 & 83.09 & 81.26 & 78.07 & 72.65 \\

OmniGuard (100)
& 99.68 & 95.85 & 84.65 & 72.55 & 62.82 \\

VisGuard (81)
& \underline{99.93}
& \underline{99.90}
& \underline{99.83}
& \underline{98.83}
& 93.28 \\

VizAnchor (81)
& \textbf{99.97}
& \textbf{99.96}
& \textbf{99.93}
& \textbf{99.87}
& \textbf{99.43} \\
\bottomrule
\end{tabular*}
\caption{Metadata BitAcc (\%) under different modified-pixel ratios on VGD.}
\label{tab:modified_pixel_recovery}
\end{table}

\begin{table}[t]
\centering
\small
\setlength{\tabcolsep}{2.6pt}
\begin{tabular*}{\columnwidth}{@{\extracolsep{\fill}}cccccc}
\toprule
\multirow[c]{2}{*}{Method}
& \multicolumn{5}{c}{Cropped Area Ratio (\%)} \\
\cmidrule(lr){2-6}
& 10 & 20 & 30 & 40 & 50 \\
\midrule
WAM (32)
& 87.47
& 87.36
& 87.14
& 86.62
& 85.49 \\

VisGuard (81)
& \underline{97.77}
& \underline{98.65}
& \underline{98.97}
& \underline{98.73}
& \underline{97.63} \\

VizAnchor (81)
& \textbf{99.55}
& \textbf{99.61}
& \textbf{99.59}
& \textbf{99.55}
& \textbf{99.36} \\
\bottomrule
\end{tabular*}
\caption{Metadata BitAcc (\%) under different cropped-area ratios on VGD.}
\label{tab:crop_metadata_recovery}
\end{table}

\subsection{Spatial Anchor Construction}
\label{subsec:spatial_anchor_eval}

We evaluate whether the Spatial Anchor accurately localizes both crop manipulations and local edits, providing reliable spatial evidence for downstream manipulation reasoning.

\paragraph{Crop Localization.}
We evaluate the Crop-Aware Module on VGD under different cropped-area ratios and the VAD-LocEval crop subset. Other tamper-localization baselines are not included because they are not designed to handle crop-based manipulations and therefore are not directly applicable to this evaluation. Table~\ref{tab:crop_localization_visguard} and
Table~\ref{tab:crop_localization_vizanchor} show that VizAnchor consistently outperforms VisGuard, achieving higher IoU across all crop ratios and an F1 of 0.9893 on the VAD-LocEval crop subset, confirming the effectiveness of the position-map-based Crop-Aware Module.

\begin{table}[t]
\centering
\small
\setlength{\tabcolsep}{3.8pt}
\begin{tabular*}{\columnwidth}{@{\extracolsep{\fill}}cccccc}
\toprule
\multirow{2}{*}{Method}
& \multicolumn{5}{c}{Cropped Area Ratio (\%)} \\
\cmidrule(lr){2-6}
& 10 & 20 & 30 & 40 & 50 \\
\midrule
VisGuard
& 0.9789
& 0.9815
& 0.9824
& 0.9814
& 0.9684 \\

VizAnchor
& \textbf{0.9866}
& \textbf{0.9875}
& \textbf{0.9860}
& \textbf{0.9837}
& \textbf{0.9790} \\
\bottomrule
\end{tabular*}
\caption{Crop localization IoU on VGD.}
\label{tab:crop_localization_visguard}
\end{table}

\begin{table}[t]
\centering
\small
\setlength{\tabcolsep}{5.0pt}
\begin{tabular*}{\columnwidth}{@{\extracolsep{\fill}}ccccc}
\toprule
Method
& IoU $\uparrow$
& Prec. $\uparrow$
& Rec. $\uparrow$
& F1 $\uparrow$ \\
\midrule
VisGuard
& 0.9710
& \textbf{0.9982}
& 0.9726
& 0.9851 \\

VizAnchor
& \textbf{0.9789}
& 0.9981
& \textbf{0.9808}
& \textbf{0.9893} \\
\bottomrule
\end{tabular*}
\caption{Crop localization on crop subset of VAD-LocEval.}
\label{tab:crop_localization_vizanchor}
\end{table}

\begin{table}[t]
\centering
\small
\setlength{\tabcolsep}{3.0pt}
\begin{tabular*}{\columnwidth}
{@{\extracolsep{\fill}}ccccc}
\toprule
Method
& IoU $\uparrow$
& Prec. $\uparrow$
& Recall $\uparrow$
& F1 $\uparrow$ \\
\midrule
EditGuard
& 0.5062
& 0.5969
& 0.7399
& 0.6289 \\

VizDefender
& 0.5324
& 0.6178
& 0.7584
& 0.6556 \\

VisGuard
& \underline{0.6067}
& \underline{0.7668}
& \underline{0.7622}
& \underline{0.7152} \\

VizAnchor
& \textbf{0.7418}
& \textbf{0.8011}
& \textbf{0.9105}
& \textbf{0.8375} \\
\bottomrule
\end{tabular*}
\caption{Overall local-edit localization results across VDD-LocEval and local-edit subset of VAD-LocEval.}
\label{tab:overall_localization}
\end{table}

\paragraph{Local-Edit Localization.}
We evaluate the Localization Module on VDD-LocEval and the local-edit subset of VAD-LocEval. OmniGuard is not included as a localization baseline because its public release does not provide the tamper localization weights, while only the watermark recovery component is available. Table~\ref{tab:overall_localization} shows that VizAnchor substantially outperforms all baselines, improving overall IoU to 0.7418 and F1 to 0.8375.

\subsection{Decoding Manipulation Intent with VLMs}
\label{subsec:vlm_decoding}

We evaluate VLM-based manipulation understanding on VDD-ReasonEval and VAD-ReasonEval across four tasks: tamper-type classification, tampered-component recognition, manipulation-process generation, and misleading-intent inference. All agents share the same gemini-3.5-flash backbone accessed through the Google Gemini API.

\begin{figure}[t]
    \centering
    \includegraphics[width=\columnwidth]{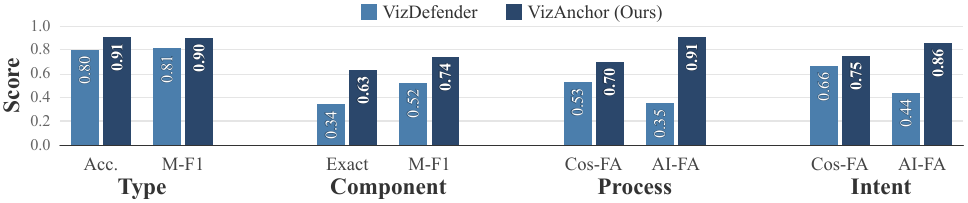}
    \caption{
    Sample-weighted tamper-type, component recognition, manipulation-process and misleading-intent results across VAD-ReasonEval and VDD-ReasonEval. Exact denotes exact-match accuracy, and M-F1 denotes Macro-F1.
    }
    \label{fig:overall_inferring}
\end{figure}

\paragraph{Tamper Type Classification.}
We formulate tamper type prediction as closed-set single-label classification over the predefined taxonomy. As shown in Figure~\ref{fig:overall_inferring}, VizAnchor achieves an Accuracy of 0.91 and a Macro-F1 of 0.90, outperforming VizDefender by 0.11 and 0.09, respectively. The comparable Accuracy and Macro-F1 indicate consistent recognition across different manipulation categories.

\paragraph{Tampered Component Recognition.}
We formulate tampered-component recognition as multi-label
classification because one manipulation may affect multiple chart
elements. Figure~\ref{fig:overall_inferring} shows that VizAnchor
improves Exact Match Accuracy from 0.34 to 0.63 and Macro-F1 from 0.52 to 0.74. These substantial gains indicate that the dual-anchor evidence helps identify both the manipulated regions and their corresponding chart components.

\paragraph{Manipulation Process and Misleading Intent.}
We evaluate the manipulation processes grounded by MGA and misleading intents  inferred by IIA. As shown in Figure~\ref{fig:overall_inferring}, VizAnchor outperforms VizDefender on both tasks. For process and intent, VizAnchor achieves Cos-FA scores of 0.70 and 0.75 (vs. 0.53 and 0.66) and AI-FA scores of 0.91 and 0.86 (vs. 0.35 and 0.44). These improvements, particularly in process evaluation, confirm that our dual-anchor prompt enables precise spatial grounding of concrete chart edits, which drives accurate inference of the misleading intent.

\subsection{Ablation Study}
\label{subsec:ablation}

\begin{figure}[t]
    \centering
    \includegraphics[width=\columnwidth]{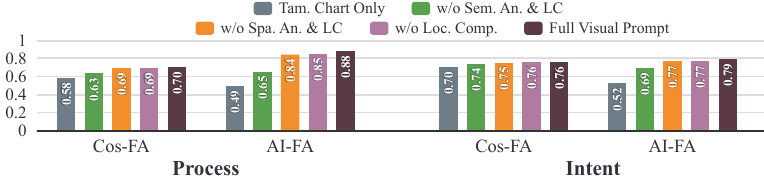}
    \caption{
    Overall visual-evidence ablation results across VAD-ReasonEval and VDD-ReasonEval.
    }
    \label{fig:overall_visual_evidence_ablation}
\end{figure}

\paragraph{Visual Evidence.}
We conduct a single-call ablation using gemini-3.5-flash, comparing the complete visual prompt with variants lacking the Spatial Anchor, Semantic Anchor, or Localized Comparison (LC), plus a tampered-chart-only baseline. As shown in
Figure~\ref{fig:overall_visual_evidence_ablation}, the Semantic Anchor provides the largest gain, raising tamper-type Macro-F1 from 0.473 to 0.778. The Spatial Anchor focuses reasoning on manipulated regions, while LC highlights fine-grained before--after differences. Together, they provide complementary semantic, spatial, and comparative evidence. This controlled single-call setting is used only for comparisons among the visual-input configurations and is not directly comparable with our full model. Detailed results are provided in the appendix.

\paragraph{Multi-Agent Reasoning.}
We evaluate the contribution of each agent to the final intent inference. The full pipeline achieves the best performance with an Intent Cos. FA of 0.753 and an AI FA of 0.856. Removing MGA or CNRA degrades performance (dropping to 0.734/0.806 and 0.743/0.832 for Cos./AI FA, respectively). Relying solely on the Intent Inferring Agent (IIA Only) yields the lowest scores (0.721 and 0.826), confirming that prior grounding and narrative reconstruction are essential for accurate intent reasoning. Detailed per-dataset metrics are provided in the appendix.

\section{User Study}
\label{subsec:user_study}

\begin{figure}[t]
    \centering
    \includegraphics[width=\columnwidth]{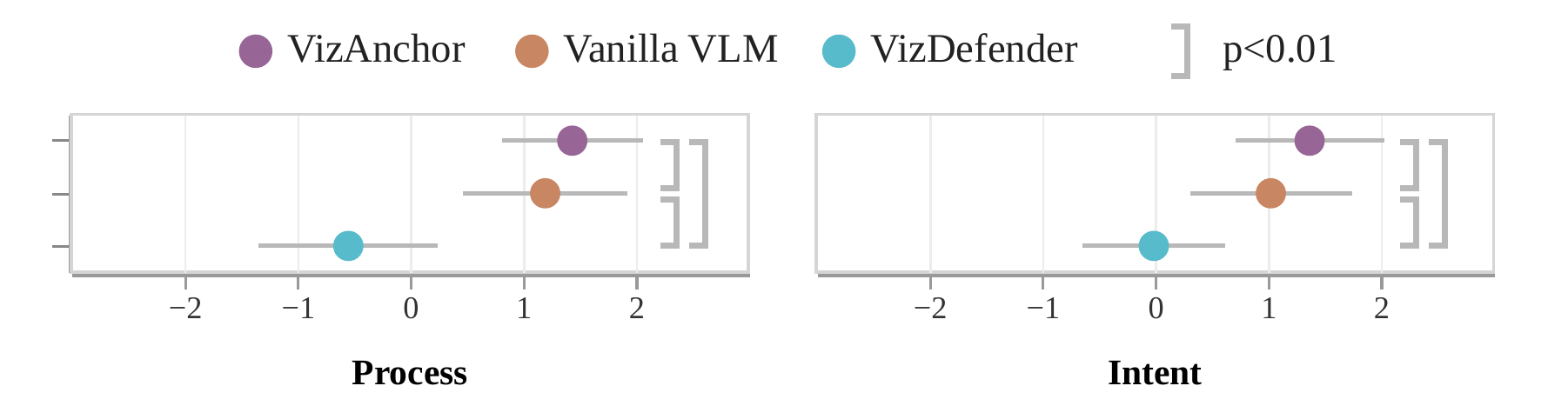}
    \caption{
    The results of the user study. Each circle represents the mean score, with error bars indicating the standard deviation. The statistical significance was assessed using the Wilcoxon signed-rank test.
    }
    \label{fig:user_study}
\end{figure}

To evaluate human-perceived explanation quality, we conducted a blind user study where 30 participants rated 10 anonymized manipulation examples on a five-point Likert scale ($\mu_{\mathrm{age}} = 24.07$ years, $\sigma = 4.28$; 16 males, 13 females, and 1 who preferred not to disclose their gender). We compared VizAnchor against VizDefender and Dual-Anchor with a vanilla VLM. As shown in Figure~\ref{fig:user_study}, VizAnchor achieved the highest average ratings across both dimensions, scoring 1.437 for manipulation process and 1.370 for misleading intent, effectively outperforming VizDefender (-0.547 and -0.010) and the vanilla VLM (1.197 and 1.027). These results validate that combining dual-anchor evidence with multi-agent reasoning yields significantly more accurate and insightful analyses of tampered visualizations.

\section{Discussion}
\label{sec:discussion}

While VizAnchor demonstrates the potential of combining proactive watermarking with multi-agent reasoning, it has two primary limitations. First, its reliance on pre-embedded metadata restricts its applicability to charts protected at the time of creation. Consequently, it cannot be directly applied to legacy or third-party visualizations already circulating online without such signals. Second, our anchors are designed to localize partial manipulations, such as cropping or localized pixel edits. With the rapid advancement of AIGC technologies, attackers can now easily regenerate an entire chart from scratch to alter its narrative. Such holistic regeneration destroys embedded signals and bypasses local forensic traces, rendering the anchors ineffective. These challenges reflect a broader shift in deceptive visualizations—from image-level localized editing to full semantic synthesis. Future work should explore integrating proactive protection with passive forensic cues, moving beyond localized evidence to verify the global semantic consistency of visualizations.

\section{Conclusion}
\label{sec:conclusion}

We present \textbf{VizAnchor}, a dual-anchor framework for visualization manipulation understanding beyond conventional tamper localization. VizAnchor constructs a Semantic Anchor from provenance-verified original charts and a Spatial Anchor from localized tamper evidence, enabling VLMs to reason over authentic and manipulated visual messages. Building upon these anchors, our multi-agent framework grounds manipulation attributes, reconstructs visualization narratives, and infers misleading intents. Extensive experiments demonstrate that VizAnchor achieves reliable metadata recovery, accurate tamper localization, and evidence-grounded explanations of manipulation tactics and misleading effects. This work highlights the potential of combining trustworthy evidence construction with multimodal reasoning for building reliable visualization communication.

\bibliography{aaai2027}

\clearpage
\onecolumn
\input{\vizanchorappendixdir/appendix_arxiv.tex}

\end{document}

%% file: VizAnchor_ArXiv_Supplement_20260825/arxiv_appendix_preamble.tex
\providecommand{\vizanchorappendixdir}{VizAnchor_ArXiv_Supplement_20260825}

\usepackage{amsfonts}
\usepackage{longtable}
\usepackage{array}
\usepackage{enumitem}
\usepackage{tcolorbox}
\tcbuselibrary{breakable,skins,listings}
\usepackage{microtype}

\definecolor{titleblockcolor}{HTML}{353535}
\definecolor{textblockcolor}{HTML}{FFFFFF}

\newenvironment{block}[2][]{%
  \begin{tcolorbox}[
    adjusted title=#2,
    fonttitle={\normalsize\bfseries},
    colback={textblockcolor},
    colframe={titleblockcolor},
    coltitle={white},
    arc=0pt,
    outer arc=0pt,
    left=5pt,
    right=5pt,
    top=5pt,
    bottom=5pt,
    fontupper=\normalsize,
    breakable,
    #1]
}{\end{tcolorbox}}

\newtcbox{\highlight}[1][red]{%
  on line,
  arc=0pt,
  outer arc=0pt,
  colback=#1!10!white,
  colframe=#1!50!black,
  boxsep=0pt,
  left=1pt,
  right=1pt,
  top=2pt,
  bottom=2pt,
  boxrule=0pt,
  bottomrule=1pt,
  toprule=1pt
}

\newcommand{\promptheading}[1]{%
  \par\medskip\noindent\textit{\textbf{#1}}\\[-2pt]
  \noindent\rule{\linewidth}{0.4pt}\par\smallskip
}

\newcommand{\promptfield}[2]{%
  \noindent\textbf{#1:} \highlight[orange]{#2}\par
}

%% file: VizAnchor_ArXiv_Supplement_20260825/appendix_arxiv.tex
\appendix
\setcounter{secnumdepth}{2}
\captionsetup{font={normalsize,rm},labelfont=normalfont,textfont=normalfont}

\section*{Appendices and Supplemental Materials}

\noindent
\begin{tabular}{@{}ll@{}}
\textbf{\ref{app:dataset_details}:} & Dataset Details and Taxonomy \\
\textbf{\ref{app:implementation_details}:} & Implementation and Inference Details \\
\textbf{\ref{app:prompts}:} & Prompts and Output Schemas \\
\textbf{\ref{app:localization_results}:} & Additional Tampering-Localization Results \\
\textbf{\ref{app:vlm_results}:} & Additional VLM Reasoning Results \\
\textbf{\ref{app:error_analysis}:} & Error Analysis \\
\textbf{\ref{app:user_study}:} & User Study Details \\
\textbf{\ref{app:ai_disclosure}:} & Disclosure of AI-Assisted Editing \\
\end{tabular}

\section{Dataset Details}
\label{app:dataset_details}

\subsection{IWM Training and Test Split}

The IWM experiments use the VisGuard dataset, which contains 17,957
chart images in total. We divide it into 14,964 training images and
2,993 held-out test images using a deterministic 5:1 random split with
seed 20260710.

\subsection{Robustness-Test Sample Construction}

We construct the modified-pixel and crop robustness samples on the fly
from all 2,993 held-out IWM test images. Each clean RGB chart is first
watermarked at the model's internal input resolution, and the resulting
watermark residual is mapped back to the chart's original spatial
resolution. The attack is then applied to this original-resolution
watermarked chart. The source-image order, payload, donor selection,
and attack geometry are generated deterministically with seed 20260713.
The source image and attack geometry for a given image--ratio pair are
therefore identical across the compared methods. Modified-pixel and crop
corruptions are evaluated separately and are never combined in these
tests.

For the modified-pixel test, let $M\in\{0,1\}^{H\times W}$ be an
irregular binary mask. We define the modified-pixel ratio as
\begin{equation}
    r_{\mathrm{mod}}=\frac{1}{HW}\sum_{u=1}^{H}\sum_{v=1}^{W}M_{uv}.
\end{equation}
For each image, we evaluate $r_{\mathrm{mod}}\in
\{0.15,0.30,0.45,0.60\}$. To generate $M$, we sample a seeded
$32\times32$ random field, bicubically upsample and spatially smooth it,
and select the $\operatorname{round}(r_{\mathrm{mod}}HW)$ pixels with
the largest field values. This produces a spatially coherent irregular
region with the requested area up to integer-pixel rounding. A
deterministically selected, different test chart is resized to $H\times
W$ and used as the donor image $I_d$. Given the watermarked chart $I_w$,
the attacked sample is
\begin{equation}
    I_{\mathrm{mod}}=(1-M)\odot I_w+M\odot I_d.
\end{equation}
Thus, only pixels selected by $M$ are replaced; no crop or additional
corruption is applied. This yields 2,993 samples at each ratio, or
11,972 modified-pixel test instances per evaluated method.

For the crop test, the reported crop ratio denotes the fraction of the
original image area that is removed, rather than the retained fraction:
\begin{equation}
    r_{\mathrm{crop}}=1-\frac{H_cW_c}{HW}.
\end{equation}
For a target $r_{\mathrm{crop}}$, we set
$s=\sqrt{1-r_{\mathrm{crop}}}$,
$H_c=\operatorname{round}(sH)$, and
$W_c=\operatorname{round}(sW)$. A seeded top-left coordinate is sampled
uniformly from all valid placements, and the corresponding
$H_c\times W_c$ rectangular region is retained. The cropped observation
is extracted directly from the original-resolution watermarked chart
without resizing or padding during sample construction. The retained
rectangle is also stored as the ground-truth crop box; any canonical-canvas restoration is performed later by the evaluated crop-recovery
pipeline. We evaluate $r_{\mathrm{crop}}\in
\{0.10,0.20,0.30,0.40,0.50\}$ and record the realized ratio after
integer rounding. This yields 2,993 samples at each ratio, or 14,965
crop test instances per evaluated method.

\subsection{Localization Data Construction and Evaluation Splits}

VAD-LocTrain contains 1,500 automatically generated local-tampering
pairs. For each sample, we select a source chart and deterministically sample one
chart-aware edit using seed 20260617 plus the sample index. Three edit
families are used. Text edits detect text-like connected components and
replace the selected local content with a different numeric or textual
token. Bar edits detect rectangular chart marks using contour
rectangularity and line cues, and shorten or extend the selected bar.
Scatter edits detect colored connected components and move a selected
point to a different location. If the requested primitive is not
detected in a chart, the generator falls back to an available edit
family. The final collection contains 685 text edits, 517 bar edits,
and 298 scatter-point edits.

Each generation operation saves the clean source chart, the tampered
chart, a binary pixel mask covering the modified source and/or target
region, a visualization overlay, and a JSONL provenance record
containing the source path, operation, coordinates, and random seed.
The masks are generated directly from the executed edit rather than
predicted by a model. All 1,500 requested samples are completed, with
no failed or unreadable outputs. The localization model uses the clean
chart, its generated tampered counterpart, and their binary RGB
difference as its seven-channel input.

The 1,500 generated pairs are shuffled and divided at the generated
sample level into 1,200 training pairs and 300 validation pairs using
seed 20260617. The validation split is used only for checkpoint
selection and hyperparameter monitoring; no results on this split are
reported as final test performance. Final localization results are
evaluated separately on the manually constructed VAD-LocEval and the
independent VDD-LocEval.

VAD-Eval contains 120 manually created chart pairs. All 120 are used
for manipulation-understanding evaluation as VAD-ReasonEval. For
localization, VAD-LocEval contains 20 crop samples for crop localization
and 100 non-crop local-edit samples for tamper-region localization.
VAD-LocEval is not used to update the localization-network parameters.

VDD-LocEval contains 1,000 samples used for local-edit localization
evaluation. VDD-ReasonEval contains 100 samples used for
manipulation-understanding evaluation.

\subsection{Reasoning-Annotation Protocol}

The reference annotations used for manipulation-understanding
evaluation were produced by two experts in data visualization. The
annotations cover the tamper type, tampered components, concrete
manipulation process, and misleading intent for each evaluation sample.
After the initial annotation, the complete set of reference answers
underwent a second verification pass to check label consistency and the
factual correspondence between each answer and its original--tampered
chart pair. The verified annotations are used only as evaluation
references and are never provided to the reasoning agents during
inference.

We define nine chart-specific tamper types: Modifying Data Point Values
(MDV), Adding or Removing Data Points (ARD), Modifying Coordinate
Values (MCV), Deceptive Auxiliary Annotations (DAA), Modifying Legend
(ML), Hiding Labels (HL), Adding or Removing Logos (ARL), Data--Visual
Disproportion (DVD), and Modifying Colormap (MC). We further define
seven tampered-component categories: Data Label, Region, Axis,
Annotation, Legend, Logo, and Colormap.
For evaluation, \texttt{mix} denotes samples containing multiple tampering
mechanisms.

\clearpage

\section{Implementation Details}
\label{app:implementation_details}

\subsection{Invertible Watermarking Module}
\label{app:iwm_details}

The metadata capacity is $K=81$ bits and is represented as an
$h_m \times w_m=9\times9$ binary map. We use tiling factors
$r_h=r_w=3$, resulting in a $27\times27$ redundant module map.
The canonical chart resolution is $H\times W=512\times512$.
The IWM uses patch size $p=16$, token dimension $d=768$, latent
dimension $d_z=64$, and $L=4$ invertible transformation blocks.
During crop-aware training, the crop side length is sampled from
$[102,512]$, the transition strength is set to $0.01$, and
\texttt{mask\_texture\_type} is randomly sampled from 0, 1, and 2,
corresponding to no mask, random mask-texture occlusion, and the
combined masking/padding case, respectively.

\subsection{IWM Training Configuration}
\label{app:training_configuration}

The IWM and Crop-Aware Module are jointly optimized using the following
weighted objective:
\begin{equation}
\begin{aligned}
\mathcal{L}_{\mathrm{IWM}}
={}&
30.0\,\mathcal{L}_{\mathrm{steg}}
+0.2\,\mathcal{L}_{\mathrm{ssim}}
+25.0\,\mathcal{L}_{\mathrm{meta}} \\
&+
2.0\,\mathcal{L}_{\mathrm{lpips}}
+10.0\,\mathcal{L}_{\mathrm{pf}}.
\end{aligned}
\end{equation}
Here, $\mathcal{L}_{\mathrm{steg}}$ is the $\ell_1$ loss between the
original chart and the watermarked chart,
$\mathcal{L}_{\mathrm{ssim}}=1-\operatorname{SSIM}$,
and $\mathcal{L}_{\mathrm{meta}}$ is the
\texttt{BCEWithLogitsLoss} applied to the predicted metadata logits.
$\mathcal{L}_{\mathrm{lpips}}$ measures the LPIPS distance between the
original and watermarked charts, while $\mathcal{L}_{\mathrm{pf}}$ is
the $\ell_1$ loss between the predicted and target position maps.

The IWM is trained for 40 epochs with a batch size of 4 on one GPU.
We use AdamW with an initial learning rate of $1\times10^{-4}$,
$\beta_1=0.9$, $\beta_2=0.999$, and weight decay $1\times10^{-5}$.
The learning rate follows a StepLR schedule and is multiplied by 0.9
after every epoch. Training uses random seed 20260701. Checkpoints are
saved after every epoch, and the reported IWM is the final epoch-40
checkpoint; no validation-based early stopping is used. The model is
trained from scratch without loading pretrained watermarking weights.
We do not apply the optional model-wide
\texttt{\_initialize\_weights} reinitialization. Layers therefore use
their constructor defaults, except for component-internal
initializations: the position-flow convolutional and linear layers use
Kaiming initialization scaled by 0.01, and invertible mixing matrices
are initialized as random orthogonal matrices.

\subsection{Localization U-Net}
\label{app:localization_unet}

The localization network receives a seven-channel tensor formed by
concatenating an authentic RGB chart, its tampered counterpart, and a
one-channel binary RGB-difference map. In VAD-LocTrain these are the
clean source and automatically edited charts described above; in the
reported end-to-end tests they are the authentic watermarked chart and
the corresponding tampered watermarked chart:
\begin{equation}
\begin{aligned}
D(x,y)&=\max_{c\in\{R,G,B\}}|I_{\mathrm{auth}}(x,y,c)
       -I_{\mathrm{tamp}}(x,y,c)|,\\
B_{\mathrm{diff}}(x,y)&=\mathbf{1}[D(x,y)>\tau_{\mathrm{diff}}],\\
X_{\mathrm{loc}}&=\operatorname{Concat}
\left(I_{\mathrm{auth}},I_{\mathrm{tamp}},B_{\mathrm{diff}}\right),
\end{aligned}
\end{equation}
where $\tau_{\mathrm{diff}}=0$ for the reported model. During training,
all three inputs are resized to $512\times512$.

The network is a four-level U-Net with encoder channel widths
$32$, $64$, $128$, and $256$. Each level contains two $3\times3$
convolutions, each followed by batch normalization and ReLU. The first
encoder block maps 7 channels to 32 channels, and the following levels
use $2\times2$ max pooling before their double-convolution blocks. The
decoder uses bilinear upsampling, concatenates the corresponding encoder
skip feature, and applies double-convolution blocks with output widths
$128$, $64$, and $32$. A final $1\times1$ convolution maps the
32-channel feature map to one tamper-mask logit channel. Because the
network is fully convolutional, the reported localization tests can be
performed at the original chart resolution even though training uses
$512\times512$ inputs.

The localization model is trained for 65 epochs with batch size 16,
AdamW, learning rate $1\times10^{-4}$, and weight decay
$1\times10^{-4}$; no learning-rate scheduler is used. The 1,500
VAD-LocTrain samples are randomly divided into 1,200 training samples
and 300 validation samples (80/20) using seed 20260617, as detailed in
the dataset-construction protocol above. The objective is
\begin{equation}
\mathcal{L}_{\mathrm{loc}}
=1.0\,\mathcal{L}_{\mathrm{BCE}}
+1.0\,\mathcal{L}_{\mathrm{Dice}}.
\end{equation}
Thus, BCE and Dice have equal external weights of 1. The BCE term is a
\texttt{BCEWithLogitsLoss} with positive-class weight 5 to account for
the sparse tampered pixels. Checkpoint selection is based on IoU measured on the 300-sample
validation split with a probability threshold of 0.5. This validation
split is used only for model selection and is distinct from
VAD-LocEval and VDD-LocEval, which are used for final evaluation.
The epoch-65 checkpoint attains the highest validation IoU in the
reported training run. At test time, the reported binary masks use
$\tau_{\mathrm{mask}}=0.9$, selected separately from
$\tau_{\mathrm{diff}}$.

\subsection{VLM Inference Protocol}
\label{app:vlm_protocol}

The complete reasoning pipeline uses three sequential, role-specific
calls to \texttt{gemini-3.5-flash}. The Misleader Grounding Agent
(MGA) predicts the tamper type, tampered component, and manipulation
process. The Chart Narrative Reconstruction Agent (CNRA) independently
reconstructs the narratives of the authentic and tampered charts.
The Intent Inferring Agent (IIA) integrates the four-panel visual
prompt with the outputs of MGA and CNRA to infer the misleading intent.
No reference tamper type, component, manipulation process, or intent is
provided to any agent during inference. Ground-truth annotations are
used only after inference for evaluation.

All VizAnchor agent calls use \texttt{gemini-3.5-flash} through the
Google Gemini API with a temperature of 0 and a maximum output length of
8,192 tokens. The visual-evidence ablation uses the same model and inference
configuration. Strong-model evaluation uses
\texttt{gemini-3.1-pro-preview}, also with a temperature of 0 and a maximum
output length of 8,192 tokens. Semantic-similarity evaluation uses
\texttt{text-embedding-3-large}. All model-visible images are stored as
lossless PNG files. The API-based reasoning and evaluation experiments
reported in this work were conducted in July 2026.

For the visual-evidence ablation, a single
\texttt{gemini-3.5-flash} call directly generates the four task outputs
under each visual-input setting. This single-call protocol is used only
for comparisons among A1--A5 and is not directly comparable with the
complete three-agent pipeline.

\clearpage

\section{Prompts and Output Schemas}
\label{app:prompts}

This section provides the complete prompts used in the VLM reasoning
pipeline and automated evaluation. Curly-brace placeholders are
replaced with sample-specific inputs during inference, while
angle-bracket expressions describe fields that the model must fill.
The outputs are constrained to JSON objects to facilitate automatic
parsing. The model is instructed not to produce Markdown formatting or
additional text outside the specified output schema.

\subsection{Prompting Protocol}
\label{app:prompting_protocol}

Each primitive panel is resized to $512\times512$ pixels before
composition. The complete visual prompt is a two-by-two PNG collage;
the four $512\times512$ panels are not downsampled into a single
$512\times512$ image. It follows the fixed order:
\begin{enumerate}
    \item Semantic Anchor: the provenance-retrieved authentic chart;
    \item Tampered Chart: the aligned manipulated chart;
    \item Tamper Map: the predicted region overlaid on the Semantic
    Anchor;
    \item Localized Comparison: enlarged authentic and tampered crops
    shown side by side.
\end{enumerate}

For the Tamper Map overlay, the already exported 8-bit mask image is
thresholded at the display value 127 before rendering. This value is
used only to binarize the mask for visual-overlay generation; it is not
the localization network's RGB-difference threshold
$\tau_{\mathrm{diff}}$ or output-probability threshold
$\tau_{\mathrm{mask}}$, and it cannot replace either one. Connected
components smaller than 20 pixels are then removed, and no mask dilation
is applied. The retained region is rendered on the Semantic Anchor using
a magenta fill (RGB 255, 0, 255) with opacity 0.3 and a two-pixel contour;
region identifiers are not drawn. Up to the three largest regions are
used for Localized Comparison, with 15\% contextual padding and a minimum
crop size of $128\times128$ pixels.

\subsection{Misleader Grounding Agent}
\label{app:prompt_mga}

MGA receives the complete four-panel visual prompt and identifies the
tamper type, tampered component, and concrete manipulation process.

\begin{block}{Misleader Grounding Agent Prompt}
\promptheading{Task Context}

You are a \textbf{visualization-forensics expert}.

\promptheading{Data Input}

The input image contains four panels in the following order:
\begin{enumerate}[leftmargin=*,nosep]
  \item \textbf{Semantic Anchor:} the authentic original chart;
  \item \textbf{Tampered Chart:} the aligned chart that may have been manipulated;
  \item \textbf{Tamper Map:} the predicted manipulated region highlighted on the authentic chart;
  \item \textbf{Localized Comparison:} enlarged original and tampered regions shown side by side.
\end{enumerate}

\promptheading{Chain-of-Thought}

Analyze only the manipulation supported by the visual evidence.

Tasks:
\begin{enumerate}[leftmargin=*,nosep]
  \item Identify the tamper type.
  \item Identify the tampered component.
  \item Describe the concrete manipulation process.
\end{enumerate}

\promptheading{Principles}

\textbf{Tamper type.} Select exactly one of the following:
\begin{itemize}[leftmargin=*,nosep]
  \item \textbf{Modifying Data Point Values (MDV)}
  \item \textbf{Adding or Removing Data Points (ARD)}
  \item \textbf{Modifying Coordinate Values (MCV)}
  \item \textbf{Deceptive Auxiliary Annotations (DAA)}
  \item \textbf{Modifying Legend (ML)}
  \item \textbf{Hiding Labels (HL)}
  \item \textbf{Adding or Removing Logos (ARL)}
  \item \textbf{Data-Visual Disproportion (DVD)}
  \item \textbf{Modifying Colormap (MC)}
  \item \textbf{Mixed Manipulation (Mix)}
\end{itemize}

Use \textbf{Mixed Manipulation} only when two or more distinct mechanisms
occur together; do not use it for a single edit affecting multiple chart
components.

\textbf{Tampered component.} Select one or more exact names from
\textit{Data Label, Region, Axis, Annotation, Legend, Logo,} and
\textit{Colormap}.

\textbf{Requirements.}
\begin{itemize}[leftmargin=*,nosep]
  \item Describe \textbf{what was changed}, not merely where it occurred.
  \item Do not infer the attacker's misleading intent in this step.
  \item Use specific chart entities, categories, values, or time ranges when visible.
  \item Do not mention unchanged chart content.
  \item Return only the required JSON object.
\end{itemize}

\promptheading{Output Format}

Return the output in this strict format:
\begin{verbatim}
{
  "tamper_type": "<tamper type>",
  "tampered_component": ["<component 1>", "<component 2>"],
  "manipulation_process": "<concise description>"
}
\end{verbatim}
\end{block}

\subsection{Chart Narrative Reconstruction Agent}
\label{app:prompt_cnra}

CNRA receives the Semantic Anchor and the aligned tampered chart and
independently reconstructs their principal narratives.

\begin{block}{Chart Narrative Reconstruction Agent Prompt}
\promptheading{Task Context}

You are the \textbf{Chart Narrative Reconstruction Agent} for chart
tampering analysis. Reconstruct the main message conveyed by each chart
separately.

\promptheading{Data Input}

You receive exactly two images in this order:
\begin{enumerate}[leftmargin=*,nosep]
  \item \textbf{Original Chart:} the authentic chart before manipulation;
  \item \textbf{Tampered Chart:} the chart after manipulation.
\end{enumerate}

\promptheading{Chain-of-Thought}

For each chart, summarize only visually supported information that matters
to a viewer's interpretation, including when applicable:
\begin{itemize}[leftmargin=*,nosep]
  \item the principal trend or direction;
  \item important comparisons between groups, categories, or time periods;
  \item rankings or which entities appear higher, lower, better, or worse;
  \item notable magnitudes, proportions, or differences;
  \item the main conclusion a reasonable viewer would draw.
\end{itemize}

\promptheading{Principles}

\begin{itemize}[leftmargin=*,nosep]
  \item Keep the two narratives distinct.
  \item Describe what each chart communicates rather than listing every graphical element.
  \item If a value or relationship is unreadable, \textbf{do not invent it}.
  \item Do not infer the manipulator's intent, motive, purpose, or desired deception.
  \item Do not use a ground-truth answer. This agent reconstructs chart messages
  only; another agent will infer intent later.
\end{itemize}

\promptheading{Output Format}

Return strict JSON only:
\small
\begin{verbatim}
{
  "sample_id": "{sample_id}",
  "original_narrative": "<concise but complete narrative conveyed by the Original Chart>",
  "tampered_narrative": "<concise but complete narrative conveyed by the Tampered Chart>",
  "confidence": 0.0
}
\end{verbatim}
\end{block}

\subsection{Intent Inferring Agent}
\label{app:prompt_iia}

IIA receives the complete four-panel visual prompt together with the
outputs of MGA and CNRA.

\begin{block}{Intent Inferring Agent Prompt}
\promptheading{Task Context}

You are the \textbf{Intent Inferring Agent} for chart tampering analysis.

\promptheading{Data Input}

You are given three complementary evidence sources:
\begin{enumerate}[leftmargin=*,nosep]
  \item \textbf{Visual evidence:} a multi-panel image containing the Semantic
  Anchor, Tampered Chart, Tamper Map overlaid on the Semantic Anchor, and
  Localized Comparison.
  \item \textbf{Local tampering evidence from the Misleader Grounding Agent:}
  the predicted tamper type, changed components, and manipulation process.
  \item \textbf{Global chart narratives from the Chart Narrative Reconstruction
  Agent:} the message conveyed by the authentic chart and the message conveyed
  by the tampered chart.
\end{enumerate}

Infer the likely misleading intent of the manipulation: the false impression,
biased interpretation, exaggerated claim, hidden difference, source confusion,
distorted trend, altered ranking, or changed conclusion that the tampering
encourages a viewer to accept.

\promptfield{Tamper type predicted by MGA}{\texttt{\{tamper\_type\}}}
\promptfield{Tampered components predicted by MGA}{\texttt{\{tampered\_component\}}}
\promptfield{Manipulation process predicted by MGA}{\texttt{\{manipulation\_process\}}}
\promptfield{Narrative conveyed by the authentic chart}{\texttt{\{original\_narrative\}}}
\promptfield{Narrative conveyed by the tampered chart}{\texttt{\{tampered\_narrative\}}}

\promptheading{Chain-of-Thought}

\begin{enumerate}[leftmargin=*,nosep]
  \item Verify the localized change using the visual input and MGA evidence.
  \item Compare the authentic narrative with the tampered narrative.
  \item Determine how the localized manipulation produces the changed chart message.
  \item Translate that message change into one clear statement of likely misleading intent.
\end{enumerate}

\promptheading{Principles}

\begin{itemize}[leftmargin=*,nosep]
  \item Use the visual input in this step; the textual evidence does not replace
  the chart comparison.
  \item Do not merely repeat the tampering process.
  \item Do not simply copy either narrative.
  \item State the affected subject, group, category, time period, source,
  value, trend, comparison, ranking, or conclusion when visually supported.
  \item Do not invent a social, political, or policy motive that is not
  supported by the chart.
  \item Output one natural-language intent description, not a short category label.
  \item Treat all MGA and CNRA fields as \textbf{model predictions}, not as
  ground-truth labels.
\end{itemize}

\promptheading{Output Format}

Return strict JSON only:
\scriptsize
\begin{verbatim}
{
  "sample_id": "{sample_id}",
  "pred_tamper_intent": "<one clear natural-language description of the likely misleading intent>",
  "intent_evidence": "<how the localized change transforms the authentic message into the tampered message>",
  "confidence": 0.0
}
\end{verbatim}
\end{block}

\subsection{Single-Call Visual-Evidence Ablation}
\label{app:prompt_visual_ablation}

For A1--A5, a single model call directly predicts all four task
outputs. Only the visual panels specified by the corresponding
ablation setting are provided. For A2, the predicted mask is overlaid on the tampered chart rather
than on the Semantic Anchor. In A4 and A5, the Tamper Map follows the
complete-prompt construction described in Sec.~\ref{app:prompting_protocol}.

\begin{block}{Single-Call Visual-Evidence Ablation Prompt}
\promptheading{Task Context}

You are analyzing chart tampering from the visible panels for neutral sample
\highlight[orange]{\texttt{\{neutral\_sample\_id\}}}.

\promptheading{Data Input}

\promptfield{Active setting}{\texttt{\{setting\_description\}}}

Use only the evidence visible in these panels. Do not assume that panels or
evidence not described above are available. Colored overlays are visual
localization aids, not chart content and not a tamper type.

\promptheading{Chain-of-Thought}

Return predictions for all four tasks.

\promptheading{Principles}

\textbf{Tamper type:} choose exactly one of these labels:
\begin{itemize}[leftmargin=*,nosep]
  \item \texttt{10mix}
  \item \texttt{1modifying\_data\_point\_values}
  \item \texttt{2adding\_or\_removing\_data\_points}
  \item \texttt{3modifying\_coordinate\_values}
  \item \texttt{4deceptive\_auxiliary\_annotations}
  \item \texttt{5modifying\_legend}
  \item \texttt{6hiding\_labels}
  \item \texttt{7adding\_or\_removing\_logos}
  \item \texttt{8data\_visual\_disproportion}
  \item \texttt{9modifying\_colormap}
\end{itemize}

\textbf{Type distinctions.}
\begin{itemize}[leftmargin=*,nosep]
  \item Use \texttt{10mix} when two or more distinct tampering mechanisms from
  the allowed labels occur together. Do not reduce a mix to only one constituent
  mechanism.
  \item Use \texttt{5modifying\_legend} only when legend text, categories,
  entries, order, or category-to-series meaning is changed.
  \item Use \texttt{9modifying\_colormap} when the data-to-color mapping,
  colorbar, continuous scale, palette, or color encoding changes. A changed
  color swatch in a legend is still
  \texttt{9modifying\_colormap} when color encoding is the misleading mechanism.
  \item Use \texttt{1modifying\_data\_point\_values} when the same data item
  remains but its numeric value or plotted position changes.
  \item Use \texttt{2adding\_or\_removing\_data\_points} when data items,
  bars, rows, categories, dates, markers, or series elements are inserted or
  absent. Do not call an absent item a changed value.
  \item Use \texttt{3modifying\_coordinate\_values} when axis scales, ticks,
  coordinate labels, or coordinate values change.
  \item Use \texttt{4deceptive\_auxiliary\_annotations} when a title, note,
  callout, guide, arrow, reference line, or other auxiliary annotation creates
  the deception.
  \item Use \texttt{6hiding\_labels} only when text labels or annotations are
  hidden while their underlying data items remain. If the data item and label
  disappear together, use
  \texttt{2adding\_or\_removing\_data\_points}.
  \item Use \texttt{7adding\_or\_removing\_logos} when a source, publisher,
  watermark, or branding logo is added or removed.
  \item Use \texttt{8data\_visual\_disproportion} when a drawn size or
  proportion is inconsistent with its stated value.
  \item If both legend semantics and color encoding change, choose the primary
  mechanism; prefer \texttt{9modifying\_colormap} when color meaning drives
  the deception.
  \item Cropping is an image operation, not an allowed tamper type and not an
  eleventh class. Determine the semantic chart content the crop removes or
  hides, then choose the corresponding allowed label. If cropping hides content
  from two or more distinct tampering mechanisms, choose \texttt{10mix}. A
  missing label alone is \texttt{6hiding\_labels}; a data item missing with its
  mark is \texttt{2adding\_or\_removing\_data\_points}; a missing logo combined
  with hidden labels or another distinct manipulation is \texttt{10mix}.
\end{itemize}

\textbf{Components:} \texttt{pred\_components} must be a JSON array containing
one or more exact names from \textit{Data Label, Region, Axis, Annotation,
Legend, Logo,} and \textit{Colormap}.

\textbf{Component distinctions.}
\begin{itemize}[leftmargin=*,nosep]
  \item \textbf{Data Label:} text or a number attached directly to a data mark,
  including bar-end, point, pie-slice, or line labels.
  \item \textbf{Region:} a data mark or data area, including bar size, line
  shape, point position, pie size, filled area, map region, bubble size, or a
  manipulated visual proportion.
  \item \textbf{Axis:} axis titles, ticks, tick labels, coordinate scales,
  axis lines, and grid-related scale marks.
  \item \textbf{Annotation:} a title, subtitle, caption, note, callout,
  reference-line label, or other narrative decoration outside core data labels,
  legends, and colorbars.
  \item \textbf{Legend:} a discrete category or series key. Text inside a
  legend is Legend, not Data Label.
  \item \textbf{Logo:} a source, publisher, watermark, or branding mark.
  \item \textbf{Colormap:} a colorbar, color scale, gradient, palette,
  heatmap, or map color encoding, including numeric colorbar ticks and ranges;
  it is not Legend.
\end{itemize}

Include both \textbf{Region} and \textbf{Data Label} when both a data mark and
its attached label change. Include multiple components only when each visibly
changes.

\textbf{Process:} \texttt{pred\_tamper\_process} must describe the concrete
visible modification, not why it was made.

\textbf{Intent:} \texttt{pred\_tamper\_intent} must state the likely misleading
purpose and the perception the edit is meant to change.

\promptheading{Output Format}

Return one strict JSON object only, with no Markdown:
\begin{verbatim}
{
  "pred_tamper_type": "<one allowed label>",
  "pred_components": ["<zero or more exact component names>"],
  "pred_tamper_process": "<concrete visible modification>",
  "pred_tamper_intent": "<misleading purpose and required perception change>",
  "visual_evidence": "<brief visual evidence>",
  "confidence": 0.0
}
\end{verbatim}
\end{block}

\subsection{Evaluation Metrics}
\label{app:metric_protocol}

Tamper type is evaluated as a single-label classification task using
accuracy and Macro-F1. The human-readable type names emitted by MGA are
mapped one-to-one to the nine dataset labels before scoring; the
additional \texttt{mix} label is used when a sample contains multiple
tampering mechanisms. Tampered components are evaluated as a
multi-label task using exact-match accuracy and class-wise Macro-F1 over
the seven predefined components. Exact match requires the complete
predicted component set to equal the reference set.

For manipulation process and misleading intent, the prediction and
reference are independently encoded by
\texttt{text-embedding-3-large}; their cosine similarity is reported as
Cos. FA. AI FA is the arithmetic mean of the integer 0--5 scores
assigned independently by \texttt{gemini-3.1-pro-preview} using the rubric below, divided
by 5 so that the reported metric lies in $[0,1]$. The
strong-model evaluator receives the corresponding visual input, a
neutral sample identifier, the reference answer, and the predicted
answer; it does not receive the method name.

\subsection{Strong-Model Evaluation Prompt}
\label{app:prompt_judge}

We use \texttt{gemini-3.1-pro-preview} to evaluate the semantic quality
of the generated manipulation processes and misleading intents with
access to the corresponding visual evidence.

\begin{block}{Strong-Model Evaluation Prompt}
\promptheading{Task Context}

You are an \textbf{impartial evaluator} of visualization-manipulation
descriptions.

\promptheading{Data Input}

\promptfield{Visual input}{\texttt{<VISUAL\_PROMPT\_IMAGE>}}
\promptfield{Reference answer}{\texttt{<REFERENCE\_ANSWER>}}
\promptfield{Predicted answer}{\texttt{<PREDICTED\_ANSWER>}}

\promptheading{Chain-of-Thought}

Evaluate the prediction according to:
\begin{enumerate}[leftmargin=*,nosep]
  \item \textbf{Correctness:} whether it captures the meaning of the reference
  and is supported by the visual evidence;
  \item \textbf{Specificity:} whether it identifies relevant entities, changes, or effects;
  \item \textbf{Consistency:} whether it avoids claims that contradict the
  reference or visual evidence.
\end{enumerate}

\promptheading{Principles}

Assign one integer score from 0 to 5:
\begin{itemize}[leftmargin=*,nosep]
  \item \textbf{0:} The prediction is irrelevant or completely incorrect.
  \item \textbf{1:} The prediction contains little relevant information and
  fails to capture the core reference meaning.
  \item \textbf{2:} The prediction is partially relevant but contains a major
  error or omits essential information.
  \item \textbf{3:} The prediction is generally correct but remains incomplete,
  vague, or insufficiently specific.
  \item \textbf{4:} The prediction is correct and specific, with only minor omissions.
  \item \textbf{5:} The prediction is fully correct, specific, complete, and
  semantically consistent with the reference.
\end{itemize}

Do not penalize wording differences or valid paraphrases. Return only a JSON
object.

\promptheading{Output Format}

\begin{verbatim}
{
  "score": <integer from 0 to 5>,
  "reason": "<brief justification>"
}
\end{verbatim}
\end{block}

Table~\ref{tab:strong_eval_rubric} summarizes the ordinal criteria used
by the strong-model evaluator. Scores of 0--2 indicate incorrect or
substantially incomplete answers, a score of 3 indicates a generally
correct but insufficiently specific answer, and scores of 4--5 indicate
specific, near-complete or fully reference-consistent answers. The same
rubric is applied independently to manipulation-process and
misleading-intent predictions.

\begin{table}[!htbp]
\centering
\small
\begin{tabular}{cp{0.84\textwidth}}
\toprule
Score & Criterion \\
\midrule
0 & The prediction is irrelevant or completely incorrect. \\
1 & The prediction contains little relevant information and fails to capture the core reference meaning. \\
2 & The prediction is partially relevant but contains a major error or omits essential information. \\
3 & The prediction is generally correct but remains incomplete, vague, or insufficiently specific. \\
4 & The prediction is correct and specific, with only minor omissions. \\
5 & The prediction is fully correct, specific, complete, and semantically consistent with the reference. \\
\bottomrule
\end{tabular}
\caption{Scoring rubric for strong-model evaluation.}
\label{tab:strong_eval_rubric}
\end{table}

\clearpage

\section{Additional Tampering-Localization Results}
\label{app:localization_results}

Tables~\ref{tab:va_mcd_localization} and~\ref{tab:vd_localization}
compare tamper-region localization on VAD-LocEval and VDD-LocEval,
respectively. VizAnchor obtains the highest IoU, recall, and F1 on both
benchmarks: its IoU/F1 scores are 0.7775/0.8609 on the local-edit subset
of VAD-LocEval and 0.7382/0.8352 on VDD-LocEval. On VAD-LocEval,
VisGuard achieves slightly higher precision than VizAnchor (0.8481 vs.
0.8303), but its substantially lower recall (0.6339 vs. 0.9312) leads
to a lower overall F1 score.

\begin{table}[!htbp]
\centering
\small
\setlength{\tabcolsep}{3.0pt}
\begin{tabular}{lcccc}
\toprule
Method
& IoU $\uparrow$
& Prec. $\uparrow$
& Recall $\uparrow$
& F1 $\uparrow$ \\
\midrule
EditGuard
& 0.6957
& 0.7944
& 0.8130
& 0.7881 \\

VizDefender
& \underline{0.7019}
& 0.7994
& \underline{0.8189}
& \underline{0.7954} \\

VisGuard
& 0.5572
& \textbf{0.8481}
& 0.6339
& 0.6756 \\

VizAnchor
& \textbf{0.7775}
& \underline{0.8303}
& \textbf{0.9312}
& \textbf{0.8609} \\
\bottomrule
\end{tabular}
\caption{Tamper-region detection results on the local-edit subset of
VAD-LocEval.}
\label{tab:va_mcd_localization}
\end{table}

\begin{table}[!htbp]
\centering
\small
\setlength{\tabcolsep}{3.0pt}
\begin{tabular}{lcccc}
\toprule
Method
& IoU $\uparrow$
& Prec. $\uparrow$
& Recall $\uparrow$
& F1 $\uparrow$ \\
\midrule
EditGuard
& 0.4872
& 0.5772
& 0.7326
& 0.6130 \\

VizDefender
& 0.5154
& 0.5996
& 0.7524
& 0.6416 \\

VisGuard
& \underline{0.6116}
& \underline{0.7587}
& \underline{0.7750}
& \underline{0.7191} \\

VizAnchor
& \textbf{0.7382}
& \textbf{0.7982}
& \textbf{0.9084}
& \textbf{0.8352} \\
\bottomrule
\end{tabular}
\caption{Tamper-region detection results on VDD-LocEval.}
\label{tab:vd_localization}
\end{table}

\subsection{Qualitative Localization Comparison}

The comparison below presents the original, watermarked, and tampered charts,
followed by the ground-truth and VizAnchor localization results, for
representative 512$\times$512 examples. The leftmost column identifies the
tamper type, while the rightmost column reports the misleading intent predicted
by the VizAnchor reasoning pipeline.

\newcommand{\loccomparisonimage}[2]{%
  \begin{minipage}[t]{0.09\textwidth}\vspace{0pt}%
    \includegraphics[width=\linewidth]{\vizanchorappendixdir/Appendix_Figures/localization/best_per_category_512/#1/#2}%
  \end{minipage}%
}
\newcommand{\loccomparisonrow}[3]{%
  \parbox[t]{0.13\textwidth}{\vspace{6.2pt}\raggedright\scriptsize\textbf{#2}} &
  \loccomparisonimage{#1}{01_clean_512.png} &
  \loccomparisonimage{#1}{02_watermarked_512.png} &
  \loccomparisonimage{#1}{03_tampered_512.png} &
  \loccomparisonimage{#1}{04_gt_mask_512.png} &
  \loccomparisonimage{#1}{05_ours_mask_t09_512.png} &
  \parbox[t]{0.27\textwidth}{\vspace{6.2pt}\raggedright\fontsize{6.2}{7.0}\selectfont #3}\\[5pt]
}

\begin{figure}[p]
\centering
\begingroup
\setlength{\tabcolsep}{2pt}
\renewcommand{\arraystretch}{1.0}
\begin{tabular}{@{}ccccccc@{}}
\parbox[c]{0.13\textwidth}{\centering\scriptsize Tamper Type} &
\tiny Original &
\tiny Watermarked &
\tiny Tampered &
\tiny Ground Truth &
\tiny VizAnchor &
\parbox[c]{0.27\textwidth}{\centering\scriptsize Predicted Intent}\\[6pt]
\loccomparisonrow
{01_1modifying_data_point_values_00009}
{Modifying data-point values}
{To exaggerate the negative perception of America's COVID-19 response in South Korea by showing a unanimous 100\% negative rating and removing any positive feedback.}
\loccomparisonrow
{02_2adding_or_removing_data_points_00013}
{Adding or removing data points}
{To hide Australia's high approval rating of 84\% regarding government handling of COVID-19, altering the comparison among the selected countries.}
\loccomparisonrow
{03_3modifying_coordinate_values_00029}
{Modifying coordinate values}
{To make the share of the population living in poverty appear significantly lower, approximately halved, for all plotted countries.}
\loccomparisonrow
{04_4deceptive_auxiliary_annotations_00031}
{Deceptive auxiliary annotations}
{To deceptively emphasize or exaggerate a rising trend in CO$_2$ emissions per capita during the early industrial period.}
\loccomparisonrow
{05_5modifying_legend_00050}
{Modifying the legend}
{To mislead viewers into believing that most landline Internet subscriptions in the listed countries are very low speed (256 kbit/s--2 Mbit/s) rather than high speed (10 Mbit/s and over).}
\loccomparisonrow
{06_6hiding_labels_00055}
{Hiding labels}
{To hide the scale of the y-axis, preventing viewers from understanding the actual magnitude and quantitative values of the COVID-19 cases and recoveries.}
\loccomparisonrow
{07_7adding_or_removing_logos_00070}
{Adding or removing logos}
{To remove the chart's branding and source attribution, hiding its origin.}
\loccomparisonrow
{08_8data_visual_disproportion_00071}
{Data visual disproportion}
{To visually downplay nuclear power generation in high-income countries, making it appear much lower than its actual value relative to other categories.}
\loccomparisonrow
{09_9modifying_colormap_00085}
{Modifying the colormap}
{To create confusion by making the legend's color encoding inconsistent with the colors used in the bar charts.}
\loccomparisonrow
{10_10mix_00091}
{Mix}
{To remove the publisher's branding and omit the data point for Cairns Airport, altering the dataset presented in the chart.}
\end{tabular}
\endgroup
\end{figure}
\FloatBarrier

\clearpage

\section{Additional VLM Reasoning Results}
\label{app:vlm_results}

Cos. FA and AI FA denote cosine-based and strong-model free-form
answer evaluation, respectively; reported AI FA values are normalized to
$[0,1]$.

\subsection{Task-Level Results}

\begin{table}[!htbp]
\centering
\small
\setlength{\tabcolsep}{5.0pt}
\begin{tabular}{llcccc}
\toprule
\multirow[c]{2}{*}{Dataset}
& \multirow[c]{2}{*}{Method}
& \multicolumn{2}{c}{Type}
& \multicolumn{2}{c}{Component} \\
\cmidrule(lr){3-4}
\cmidrule(lr){5-6}
&
& Acc. $\uparrow$
& M-F1 $\uparrow$
& Exact $\uparrow$
& M-F1 $\uparrow$ \\
\midrule
VAD & VizDefender
& 0.8250
& 0.8475
& 0.2667
& 0.4216 \\

VAD & VizAnchor
& \textbf{0.9083}
& \textbf{0.9056}
& \textbf{0.5667}
& \textbf{0.6875} \\
\midrule
VDD & VizDefender
& 0.7700
& 0.7667
& 0.4200
& 0.6332 \\

VDD & VizAnchor
& \textbf{0.9000}
& \textbf{0.8985}
& \textbf{0.7000}
& \textbf{0.8075} \\
\bottomrule
\end{tabular}
\caption{Tamper-type and component-recognition results. Exact denotes
exact-match accuracy, and M-F1 denotes Macro-F1.}
\label{tab:type_component_results}
\end{table}

\begin{table}[!htbp]
\centering
\small
\setlength{\tabcolsep}{5.0pt}
\begin{tabular}{llcccc}
\toprule
\multirow[c]{2}{*}{Dataset}
& \multirow[c]{2}{*}{Method}
& \multicolumn{2}{c}{Process}
& \multicolumn{2}{c}{Intent} \\
\cmidrule(lr){3-4}
\cmidrule(lr){5-6}
&
& Cos. FA $\uparrow$
& AI FA $\uparrow$
& Cos. FA $\uparrow$
& AI FA $\uparrow$ \\
\midrule
VAD & VizDefender
& 0.5227
& 0.3317
& 0.6636
& 0.3850 \\

VAD & VizAnchor
& \textbf{0.7300}
& \textbf{0.9633}
& \textbf{0.7856}
& \textbf{0.8867} \\
\midrule
VDD & VizDefender
& 0.5422
& 0.3620
& 0.6532
& 0.5020 \\

VDD & VizAnchor
& \textbf{0.6658}
& \textbf{0.8340}
& \textbf{0.7128}
& \textbf{0.8170} \\
\bottomrule
\end{tabular}
\caption{Manipulation-process and misleading-intent evaluation
results. AI FA is normalized to $[0,1]$.}
\label{tab:process_intent_results}
\end{table}

\begin{table}[!htbp]
\centering
\small
\setlength{\tabcolsep}{5.0pt}
\begin{tabular}{lcccc}
\toprule
\multirow[c]{2}{*}{Method}
& \multicolumn{2}{c}{Type}
& \multicolumn{2}{c}{Component} \\
\cmidrule(lr){2-3}
\cmidrule(lr){4-5}
& Acc. $\uparrow$
& M-F1 $\uparrow$
& Exact $\uparrow$
& M-F1 $\uparrow$ \\
\midrule
VizDefender
& 0.8000
& 0.8108
& 0.3364
& 0.5178 \\

VizAnchor
& \textbf{0.9045}
& \textbf{0.9024}
& \textbf{0.6273}
& \textbf{0.7420} \\
\bottomrule
\end{tabular}
\caption{Sample-weighted tamper-type and component-recognition results
across VAD-ReasonEval and VDD-ReasonEval. Exact denotes exact-match
accuracy, and M-F1 denotes Macro-F1.}
\label{tab:overall_type_component_results}
\end{table}

\begin{table}[!htbp]
\centering
\small
\setlength{\tabcolsep}{5.0pt}
\begin{tabular}{lcccc}
\toprule
\multirow[c]{2}{*}{Method}
& \multicolumn{2}{c}{Process}
& \multicolumn{2}{c}{Intent} \\
\cmidrule(lr){2-3}
\cmidrule(lr){4-5}
& Cos. FA $\uparrow$
& AI FA $\uparrow$
& Cos. FA $\uparrow$
& AI FA $\uparrow$ \\
\midrule
VizDefender
& 0.5316
& 0.3468
& 0.6589
& 0.4382 \\

VizAnchor
& \textbf{0.7008}
& \textbf{0.9045}
& \textbf{0.7525}
& \textbf{0.8550} \\
\bottomrule
\end{tabular}
\caption{Sample-weighted manipulation-process and misleading-intent
evaluation results across VAD-ReasonEval and VDD-ReasonEval. AI FA is
normalized to $[0,1]$ in this aggregate table.}
\label{tab:overall_process_intent_results}
\end{table}

\subsection{Qualitative Comparisons}

Figures~\ref{fig:qualitative_00002}--\ref{fig:qualitative_00092}
compare the manipulation-process and misleading-intent descriptions produced
by VizDefender and VizAnchor. We select one case from each of the ten tamper
types. For each case, the four visual panels show the authentic chart, tampered chart, tamper-region
overlay, and localized original--tampered comparison, followed by the
expert-verified ground truth and the outputs of the two methods. Within each
output, only the key incorrect VizDefender claims are boldfaced, while the
most discriminative correct VizAnchor phrases are shown in bold red.

\input{\vizanchorappendixdir/qualitative_intent_cases_arxiv.tex}

\clearpage
\subsection{Visual-Evidence Ablation}

\begin{table}[!htbp]
\centering
\small
\begin{tabular}{lp{0.28\textwidth}p{0.52\textwidth}}
\toprule
Setting & Input & Purpose \\
\midrule
A1 & Tampered chart only
& Tests how much the VLM can infer from the final manipulated chart alone. \\

A2 & Tampered chart and mask-overlaid tampered chart
& Tests whether localization evidence helps the VLM attend to suspicious regions. \\

A3 & Semantic Anchor and tampered chart
& Tests whether authentic--tampered comparison helps identify semantic deviations. \\

A4 & Semantic Anchor, tampered chart, and Tamper Map
& Tests whether authentic-chart comparison and localization evidence are complementary. \\

A5 & Complete four-panel visual prompt
& Further includes the Localized Comparison to test fine-grained manipulation understanding. \\
\bottomrule
\end{tabular}
\caption{Visual-input settings used in the evidence ablation.}
\label{tab:input_ablation_settings}
\end{table}

\begin{table}[!htbp]
\centering
\small
\setlength{\tabcolsep}{4.5pt}
\begin{tabular}{llcccccccc}
\toprule
\textbf{Dataset} & \textbf{Setting}
& \multicolumn{2}{c}{\textbf{Tamper Type}}
& \multicolumn{2}{c}{\textbf{Component}}
& \multicolumn{2}{c}{\textbf{Process}}
& \multicolumn{2}{c}{\textbf{Intent}} \\
&
& Acc. $\uparrow$
& Macro-F1 $\uparrow$
& Exact $\uparrow$
& Macro-F1 $\uparrow$
& Cosine $\uparrow$
& AI $\uparrow$
& Cosine $\uparrow$
& AI $\uparrow$ \\
\midrule
VAD & A1
& 0.403
& 0.388
& 0.387
& 0.460
& 0.578
& 0.446
& 0.716
& 0.478 \\

VAD & A2
& 0.600
& 0.598
& 0.525
& 0.608
& 0.651
& 0.672
& 0.763
& 0.684 \\

VAD & A3
& 0.767
& 0.753
& 0.625
& \textbf{0.706}
& 0.718
& 0.892
& 0.779
& 0.806 \\

VAD & A4
& \textbf{0.775}
& \textbf{0.757}
& 0.575
& 0.683
& 0.721
& 0.926
& \textbf{0.796}
& 0.806 \\

VAD & A5
& \textbf{0.775}
& 0.752
& \textbf{0.667}
& 0.694
& \textbf{0.733}
& \textbf{0.938}
& 0.794
& \textbf{0.826} \\
\midrule
VDD & A1
& 0.576
& 0.567
& 0.525
& 0.638
& 0.574
& 0.534
& 0.681
& 0.564 \\

VDD & A2
& 0.747
& 0.748
& 0.677
& 0.765
& 0.620
& 0.634
& 0.710
& 0.698 \\

VDD & A3
& 0.798
& 0.798
& 0.697
& 0.795
& 0.650
& 0.776
& 0.706
& 0.726 \\

VDD & A4
& 0.800
& 0.801
& \textbf{0.720}
& 0.788
& 0.662
& 0.752
& \textbf{0.718}
& 0.722 \\

VDD & A5
& \textbf{0.830}
& \textbf{0.832}
& 0.680
& \textbf{0.803}
& \textbf{0.668}
& \textbf{0.804}
& 0.713
& \textbf{0.748} \\
\bottomrule
\end{tabular}
\caption{Ablation of the visual evidence used for manipulation
understanding on VAD and VDD. AI scores are normalized to $[0,1]$. The best
result for each dataset and metric is shown in bold.}
\label{tab:visual_evidence_ablation}
\end{table}

\begin{table}[!htbp]
\centering
\small
\setlength{\tabcolsep}{4.2pt}
\begin{tabular}{lcccccccc}
\toprule
\textbf{Setting}
& \multicolumn{2}{c}{\textbf{Type}}
& \multicolumn{2}{c}{\textbf{Component}}
& \multicolumn{2}{c}{\textbf{Process}}
& \multicolumn{2}{c}{\textbf{Intent}} \\
\cmidrule(lr){2-3}
\cmidrule(lr){4-5}
\cmidrule(lr){6-7}
\cmidrule(lr){8-9}

& Acc. $\uparrow$
& Macro-F1 $\uparrow$
& Exact $\uparrow$
& Macro-F1 $\uparrow$
& Cos. FA $\uparrow$
& AI FA $\uparrow$
& Cos. FA $\uparrow$
& AI FA $\uparrow$ \\

\midrule
A1
& 0.4816
& 0.4728
& 0.4495
& 0.5417
& 0.5762
& 0.4860
& 0.7001
& 0.5171 \\

A2
& 0.6668
& 0.6675
& 0.5936
& 0.6784
& 0.6369
& 0.6547
& 0.7389
& 0.6904 \\

A3
& 0.7811
& 0.7775
& 0.6575
& \textbf{0.7468}
& 0.6871
& 0.8393
& 0.7458
& 0.7696 \\

A4
& 0.7864
& 0.7813
& 0.6409
& 0.7274
& 0.6942
& 0.8469
& \textbf{0.7605}
& 0.7678 \\

A5
& \textbf{0.8000}
& \textbf{0.7953}
& \textbf{0.6727}
& 0.7446
& \textbf{0.7035}
& \textbf{0.8771}
& 0.7572
& \textbf{0.7905} \\
\bottomrule
\end{tabular}
\caption{Overall visual-evidence ablation results across
VAD-ReasonEval and VDD-ReasonEval. AI FA scores are
normalized to $[0,1]$. The best result for each metric is shown in bold.}
\label{tab:overall_visual_evidence_ablation}
\end{table}

To examine performance across manipulation and visualization forms, we
stratify the complete VizDefender and VizAnchor pipeline results by the
annotated tamper type and by chart type.  Tamper types are taken directly
from the dataset annotations.  In particular, cropping is treated only
as an image-construction operation: each crop-derived VAD sample retains
its annotated semantic tamper type and \emph{crop} is never introduced as
a class.  Chart types are assigned by auditing the original charts under
a common taxonomy.  Grouped, stacked, and diverging bars are included in
Bar; radial partitions are included in Pie/Donut; and tables,
infographics, pictograms, Sankey diagrams, heatmaps, and multi-chart
compositions are grouped as Other/Mixed.

Tables~\ref{tab:main_tamper_complete} and
\ref{tab:main_chart_complete} report the complete VizDefender and
VizAnchor pipeline results under the same grouping protocol.  For a row keyed
by ground-truth tamper type, Type Acc. is that class's recall and Type
F1 is its one-vs-rest F1 computed over the full method--dataset pool.
For a row keyed by chart type, Type Acc. and Type F1 are the accuracy
and Macro-F1 within that chart subgroup.  Component F1 is the Macro-F1 over the seven component labels. AI scores
are normalized to $[0,1]$.

\input{\vizanchorappendixdir/complete_main_category_longtables.tex}

\clearpage
\subsection{Multi-Agent Ablation}

B1 uses IIA alone to predict the misleading intent from the complete
visual prompt. B2 removes CNRA, B3 removes MGA, and B4 uses the complete
MGA--CNRA--\allowbreak IIA pipeline.

\begin{table}[!htbp]
\centering
\small
\setlength{\tabcolsep}{8.0pt}
\begin{tabular}{llcc}
\toprule
Dataset & Setting
& Intent Cosine $\uparrow$
& Intent AI $\uparrow$ \\
\midrule
VAD
& B1 IIA Only
& 0.7459
& 0.844 \\

VAD
& B2 w/o CNRA
& 0.7726
& 0.878 \\

VAD
& B3 w/o MGA
& 0.7582
& 0.860 \\

VAD
& B4 Full
& \textbf{0.7856}
& \textbf{0.886} \\
\midrule
VDD
& B1 IIA Only
& 0.6906
& 0.806 \\

VDD
& B2 w/o CNRA
& 0.7083
& 0.778 \\

VDD
& B3 w/o MGA
& 0.7042
& 0.740 \\

VDD
& B4 Full
& \textbf{0.7128}
& \textbf{0.818} \\
\bottomrule
\end{tabular}
\caption{Ablation of the multi-agent reasoning pipeline for
misleading-intent inference on VAD and VDD. Intent AI denotes the
strong-model evaluation score normalized to $[0,1]$.}
\label{tab:reasoning_ablation}
\end{table}

\begin{table}[!htbp]
\centering
\small
\setlength{\tabcolsep}{8.0pt}
\begin{tabular}{lcc}
\toprule
\textbf{Setting}
& \textbf{Intent Cos. FA} $\uparrow$
& \textbf{Intent AI FA} $\uparrow$ \\
\midrule
B1 IIA Only
& 0.7208
& 0.826 \\

B2 w/o CNRA
& 0.7434
& 0.832 \\

B3 w/o MGA
& 0.7337
& 0.806 \\

B4 Full
& \textbf{0.7525}
& \textbf{0.856} \\
\bottomrule
\end{tabular}
\caption{Sample-weighted overall ablation results for the multi-agent
reasoning pipeline across VAD-ReasonEval and VDD-ReasonEval. Intent
AI FA is normalized to $[0,1]$.}
\label{tab:overall_reasoning_ablation}
\end{table}

\clearpage
\section{Error Analysis}
\label{app:error_analysis}

\subsection{Category-Level Failure Modes}

Category-level results show that the remaining errors are concentrated
in manipulations that require connecting a local visual edit to the
chart's global encoding. VizDefender attains a tamper-type recall of
only 0.471 on Mix and 0.600 on modifying data-point values. Its
component exact-match accuracy further decreases to 0.088 on Mix and
0.100 on both modifying colormap and data--visual disproportion. These
results indicate that identifying a suspicious region does not
necessarily reveal all affected chart components or the semantic role
of the edit.

VizAnchor improves most category-level results, although modifying
colormap remains its most difficult tamper-type category, with a recall
of 0.650. Component exact-match accuracy is also relatively low for
adding or removing data points (0.292), modifying colormap (0.350), and
Mix (0.529). This difference between tamper-type recall and component
exact match partly reflects the strictness of the multi-label metric:
an otherwise plausible prediction is counted as incorrect when any
affected component is omitted or an additional component is predicted.
This issue is particularly pronounced for mixed and multi-region
manipulations, and may also reflect ambiguity among visually related
components such as legends, colormaps, and data marks.

\subsection{Intent-Relation Errors and Metric Disagreement}

At the category-averaged level, VizAnchor obtains intent cosine
similarities ranging from 0.691 to 0.816 and normalized AI scores
ranging from 0.700 to 0.946. Its highest cosine similarity is observed
for data--visual disproportion (0.816), while its highest AI score is
obtained for hiding labels (0.946). These manipulations often produce
comparatively explicit changes in emphasis or omitted context. The
lowest results occur for modifying colormap (0.691 cosine and 0.700
AI), adding or removing data points (0.750 AI), and Mix (0.776 AI).
These categories frequently require the model to infer how a visual
edit changes a relation involving a specific entity, comparison, or
direction. Localizing the edit is therefore necessary but not always
sufficient for recovering its misleading intent.

At the individual-example level, Fig.~11 illustrates a
compound-relation error with a normalized AI score of 0.20. The mixed
manipulation changes the map title from urban to rural population share
and simultaneously recolors the United States and China. VizAnchor
correctly identifies the urban-to-rural semantic shift and both
affected countries, but predicts an incorrect direction for the
cross-country relation specified in the reference answer. The
substantial overlap in entities and topic words likely contributes to
the relatively high cosine similarity of 0.834, whereas the incorrect
directional relation is reflected in the much lower AI score.

This example illustrates the complementary roles of the two semantic
metrics. Cosine similarity captures broad topical and lexical
alignment, whereas the AI judge is more sensitive to relation
direction, entity attribution, and contradiction. VizDefender shows
particularly low intent results for modifying colormap
(0.584 cosine and 0.180 AI) and modifying coordinate values
(0.602 cosine and 0.320 AI). Data--visual disproportion presents a
similar disagreement, with a cosine similarity of 0.761 but an AI
score of only 0.350. In these cases, the generated explanation often
identifies the general topic of the manipulation while assigning an
incorrect magnitude, direction, or entity. This pattern is consistent
with the dual anchors providing more explicit evidence for comparing
the authentic and tampered chart messages.

\subsection{Chart-Dependent Behavior}

VizAnchor's intent cosine similarity remains stable on the two largest
chart groups, reaching 0.769 on bar charts and 0.770 on line charts,
but decreases to 0.687 on pie and donut charts. Component Macro-F1 is
highest on area charts (0.964), followed by scatter plots (0.884), and
lowest on pie and donut charts (0.480). Radial encodings combine color,
angle, area, and labels within a compact region, which can make both
component assignment and the direction of the resulting message change
less explicit.

These chart-level results should be interpreted together with subgroup
size. The pie and donut group contains only eight samples, compared
with 99 bar-chart and 32 line-chart samples. Consequently, the reported
differences may reflect both chart-dependent difficulty and sampling
variation, and should not be interpreted as definitive rankings among
chart types.

\subsection{Representative Intent Error Visualizations}

\begin{figure}[!htbp]
\centering
\includegraphics[width=0.68\textwidth,height=0.60\textheight,keepaspectratio]{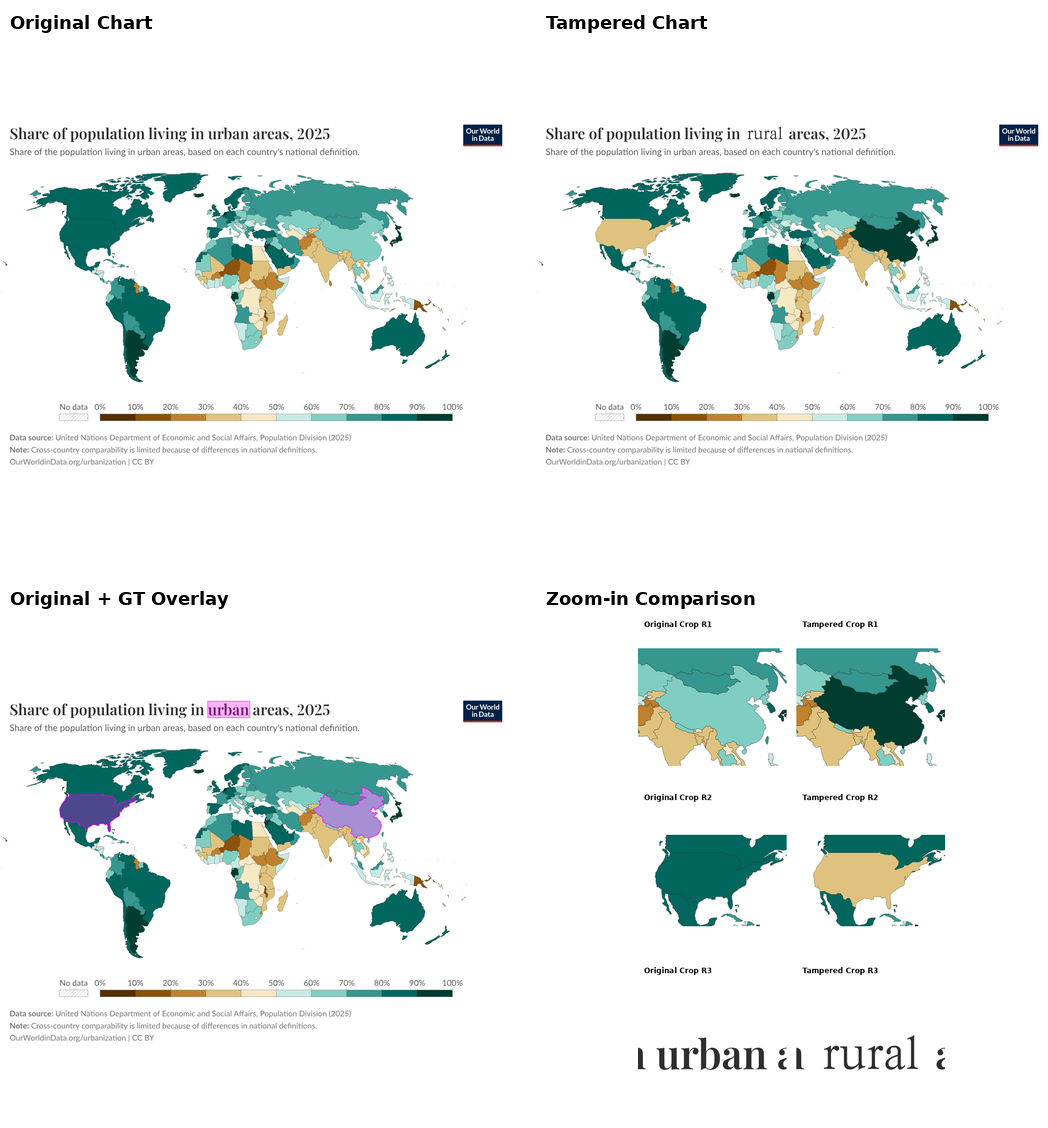}
\caption{Compound-relation error for VAD sample 00100. The title changes
from urban to rural population share while the United States and China are
also recolored. VizAnchor recognizes the altered topic and both affected
countries but does not match the benchmark reference's intended direction
of the cross-country urbanization comparison. Intent cosine remains 0.834,
while the normalized AI score is 0.20.}
\label{fig:error_vad_00100}
\end{figure}

Figure~\ref{fig:error_vad_00100} visualizes the compound-relation error discussed above.
For clarity of diagnosis, the predicted Tamper Map is replaced with
the Ground-Truth Overlay in this figure; the ground-truth mask was
not provided to the model during inference.

\noindent\textbf{Ground-truth intent.}
``To mislead viewers into thinking the map shows rural population shares
rather than urban population shares, while also making the United States
appear much less urbanized and China appear among the most urbanized
countries, thereby distorting the perceived global pattern of
urbanization.''

\noindent\textbf{VizAnchor-predicted intent.}
``To falsely depict China as having an extremely high rural population
share (80\% to 100\%) and the United States as having a much lower rural
population share (30\% to 40\%) in 2025, while misleadingly presenting a
global map of urbanization rates as representing rural population
shares.''

\noindent\textbf{Why it is incorrect.}
Both descriptions identify the urban-to-rural title substitution and the
United States and China as the affected countries. However, the predicted
relation implies that China has a low urban share and the United States has
a high urban share because rural and urban shares are complementary. The
ground-truth intent requires the opposite comparison: the United States is
made to appear less urbanized and China more urbanized. The shared entities
and topic terminology produce a high cosine similarity of 0.834, but the
reversed cross-country relation is a major semantic error, resulting in an
AI score of 0.20.

\FloatBarrier

\clearpage

\section{User Study Details}
\label{app:user_study}

\subsection{Study Design}

The user study evaluates whether human judgments support the relative
quality differences measured by the automated evaluator for two
free-form tasks: manipulation-process description and misleading-intent
inference. We selected ten chart-manipulation examples using stratified
sampling, with one example from each of the ten tamper categories and an
equal split of five examples from VAD-ReasonEval and five from
VDD-ReasonEval. Each example was accompanied by the original chart,
tampered chart, and tamper-region overlay.

For every example, participants evaluated outputs from three methods:
VizDefender, a vanilla VLM supplied with original and tampered charts, and the complete VizAnchor pipeline. The method identities were
hidden and outputs were identified only by neutral candidate numbers.
The reference answer and candidate answer were displayed in both
English and Chinese. Each participant completed 60 judgments: three
candidate outputs for each of ten examples and for each of the two
tasks. For a process answer, participants assessed whether it correctly,
specifically, and completely described the concrete visual change. For
an intent answer, they assessed whether it correctly and specifically
captured the misleading interpretation encouraged by that change.

Responses were collected on a five-point agreement scale and mapped to
$\{-2,-1,0,1,2\}$, corresponding to strongly disagree, disagree,
neutral, agree, and strongly agree. The method names were never shown to participants.
For analysis, the ten ratings for a given task--method pair were first
averaged within each participant. Thus, the participant, rather than an
individual chart judgment, is the statistical unit.

\subsection{Statistical Analysis}

We report the participant-level mean, standard deviation, and 95\%
confidence interval for each task and method. Confidence intervals use
the Student-$t$ distribution over participant-level means. A Friedman
test first evaluates the omnibus difference among the three paired
methods for each task. Pairwise comparisons use two-sided Wilcoxon
signed-rank tests; zero differences are excluded, average ranks are used
for ties, and the large-sample normal approximation is applied without
continuity correction. The three pairwise comparisons within each task
are corrected using the Holm procedure. We additionally report the
rank-biserial correlation $r_{\mathrm{rb}}$ as an effect size.

\subsection{Results}

Participants included 16 men, 13 women, and one individual who
preferred not to disclose gender. Valid age information was available
for 29 participants, with a mean of 24.07 years
($\mathrm{SD}=4.28$), a median of 22 years, and a range of 18--35
years. Their research backgrounds included data visualization or visual
analytics (13), computer vision (4), human--computer interaction (3),
education (2), design or art (1), chemistry (1), and other areas (6).
No Likert-scale responses were missing.

\begin{table}[!htbp]
\centering
\small
\setlength{\tabcolsep}{5.5pt}
\begin{tabular}{llccc}
\toprule
Task & Method & Mean & SD & 95\% CI \\
\midrule
Process & VizDefender & $-0.547$ & 0.795 & $[-0.844,-0.250]$ \\
Process & Vanilla VLM & 1.197 & 0.726 & $[0.926,1.468]$ \\
Process & VizAnchor & \textbf{1.437} & 0.624 & $[1.203,1.670]$ \\
\midrule
Intent & VizDefender & $-0.010$ & 0.632 & $[-0.246,0.226]$ \\
Intent & Vanilla VLM & 1.027 & 0.717 & $[0.759,1.294]$ \\
Intent & VizAnchor & \textbf{1.370} & 0.659 & $[1.124,1.616]$ \\
\bottomrule
\end{tabular}
\caption{Participant-level descriptive statistics. Scores
range from $-2$ to 2.}
\label{tab:user_study_descriptive}
\end{table}

The Friedman tests indicated differences
among methods for both process
($\chi^2(2)=42.660$, $p=5.45\times10^{-10}$,
Kendall's $W=0.711$) and intent
($\chi^2(2)=39.138$, $p=3.17\times10^{-9}$,
Kendall's $W=0.652$). The pairwise results are reported in
Table~\ref{tab:user_study_wilcoxon}. VizAnchor received higher ratings
than both comparison methods for both tasks.

\begin{table}[!htbp]
\centering
\small
\setlength{\tabcolsep}{4.5pt}
\begin{tabular}{llrrr}
\toprule
Task & Comparison & $W$ & Holm $p$ & $r_{\mathrm{rb}}$ \\
\midrule
Process & VizAnchor vs. VizDefender & 1.0 & $1.26\times10^{-5}$ & 0.995 \\
Process & VizAnchor vs. Vanilla VLM & 25.0 & 0.00472 & 0.737 \\
Process & Vanilla VLM vs. VizDefender & 5.0 & $1.29\times10^{-5}$ & 0.975 \\
Intent & VizAnchor vs. VizDefender & 0.0 & $1.12\times10^{-5}$ & 1.000 \\
Intent & VizAnchor vs. Vanilla VLM & 45.5 & 0.00477 & 0.670 \\
Intent & Vanilla VLM vs. VizDefender & 11.0 & $1.58\times10^{-5}$ & 0.949 \\
\bottomrule
\end{tabular}
\caption{Two-sided paired Wilcoxon signed-rank tests. Holm correction is applied across
the three pairwise comparisons within each task.}
\label{tab:user_study_wilcoxon}
\end{table}

\section{Disclosure of AI-Assisted Editing}
\label{app:ai_disclosure}

A generative AI assistant was used for language editing and
\LaTeX{} consistency checking. The authors verified all technical
content, experimental settings, analyses, and reported results and take
full responsibility for the manuscript.

%% file: VizAnchor_ArXiv_Supplement_20260825/qualitative_intent_cases_arxiv.tex
\newcommand{\vdbad}[1]{\textbf{#1}}
\newcommand{\vagood}[1]{\textcolor{red}{\textbf{#1}}}
\newcommand{\intentcase}[9]{%
\begin{figure}[p]
\centering
\begin{minipage}[t]{0.48\textwidth}
\centering
\includegraphics[width=\linewidth,height=0.19\textheight,keepaspectratio]{\vizanchorappendixdir/Appendix_Figures/qualitative/vad_#1_original.png}
\par\smallskip\textbf{Original Chart}
\end{minipage}
\hfill
\begin{minipage}[t]{0.48\textwidth}
\centering
\includegraphics[width=\linewidth,height=0.19\textheight,keepaspectratio]{\vizanchorappendixdir/Appendix_Figures/qualitative/vad_#1_tampered.png}
\par\smallskip\textbf{Tampered Chart}
\end{minipage}

\vspace{4pt}
\begin{minipage}[t]{0.48\textwidth}
\centering
\includegraphics[width=\linewidth,height=0.19\textheight,keepaspectratio]{\vizanchorappendixdir/Appendix_Figures/qualitative/vad_#1_overlay.png}
\par\smallskip\textbf{Tamper-Region Overlay}
\end{minipage}
\hfill
\begin{minipage}[t]{0.48\textwidth}
\centering
\includegraphics[width=\linewidth,height=0.19\textheight,keepaspectratio]{\vizanchorappendixdir/Appendix_Figures/qualitative/vad_#1_zoom.png}
\par\smallskip\textbf{Localized Comparison}
\end{minipage}

\vspace{5pt}
{\scriptsize
\renewcommand{\arraystretch}{1.10}
\begin{tabular}{@{}p{0.14\textwidth}p{0.80\textwidth}@{}}
\toprule
\textbf{Model} & \textbf{Process and Intent} \\
\midrule
\textbf{GT}
& \textbf{Process:} #3
\newline \textbf{Intent:} #4 \\
\textbf{VizDefender}
& \textbf{Process:} #5
\newline \textbf{Intent:} #6 \\
\textbf{VizAnchor}
& \textbf{Process:} #7
\newline \textbf{Intent:} #8 \\
\bottomrule
\end{tabular}}
\caption{Qualitative comparison for VAD sample #1. Tamper type: #2. Normalized AI scores (Process/Intent): #9.}
\label{fig:qualitative_#1}
\end{figure}
}

\intentcase
{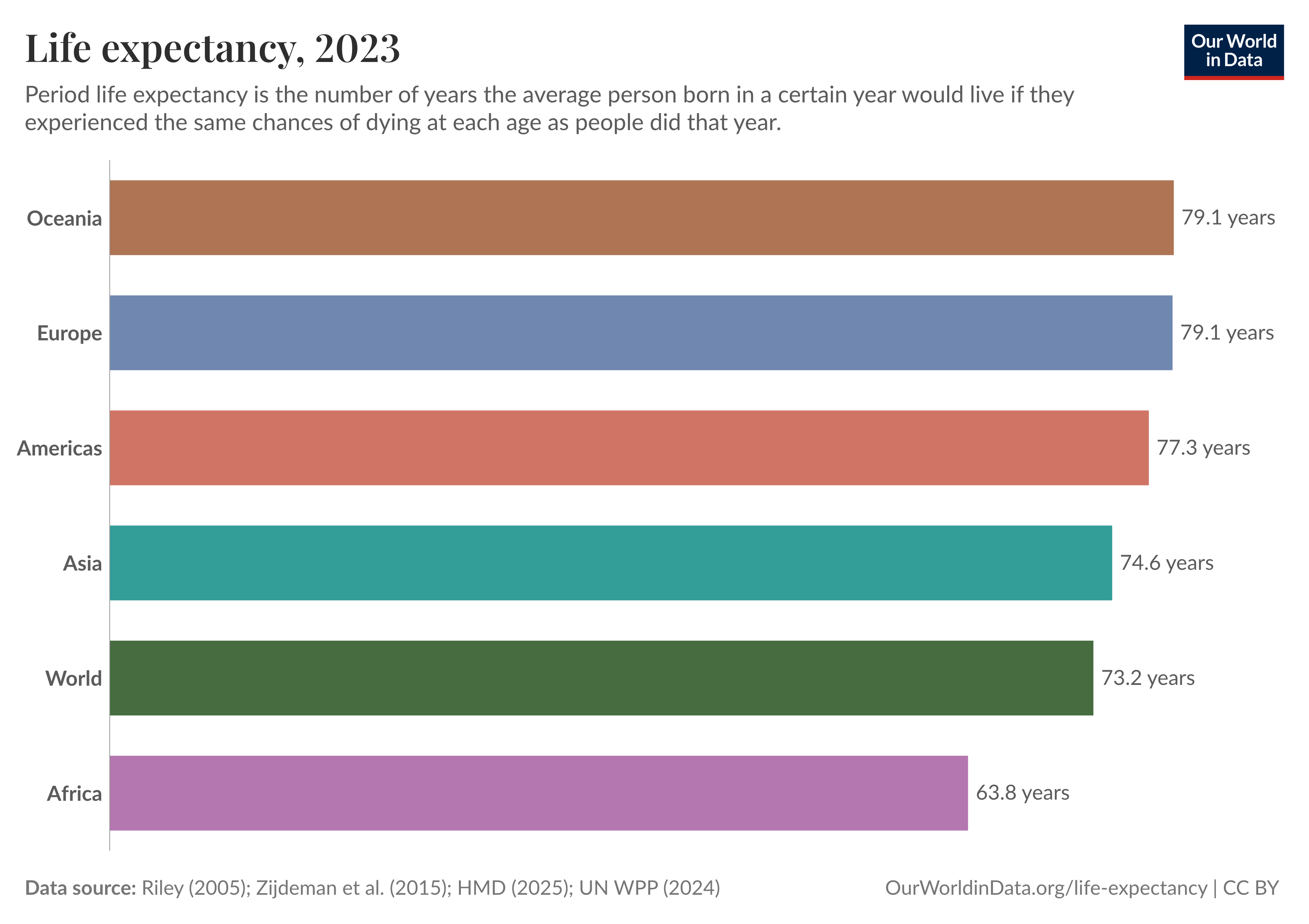}
{modifying data-point values}
{Increase Africa's life expectancy value from 63.8 years to 67.8 years and extend the corresponding bar length to match the modified value.}
{To make Africa's life expectancy appear higher than it actually is, thereby narrowing the perceived gap between Africa and other regions and downplaying regional disparities in health and living conditions.}
{The 'Africa' bar has been \vdbad{visually extended to appear longer than its actual data value of '67.8 years' would suggest}, creating a \vdbad{discrepancy between the visual representation and the numerical label}.}
{To mislead viewers into believing that the life expectancy in Africa is higher than its actual value, by making its bar \vdbad{visually disproportionate} to the provided numerical data label and other bars.}
{The bar representing 'Africa' was \vagood{extended to make it longer}, and its corresponding text label was modified \vagood{from '63.8 years' to '67.8 years'}.}
{To falsely exaggerate the life expectancy in Africa for the year 2023, making it appear \vagood{higher than it actually is} and thereby \vagood{minimizing the gap} between Africa's life expectancy and the global average.}
{VizAnchor 1.00/1.00; VizDefender 0.00/0.80}
\intentcase
{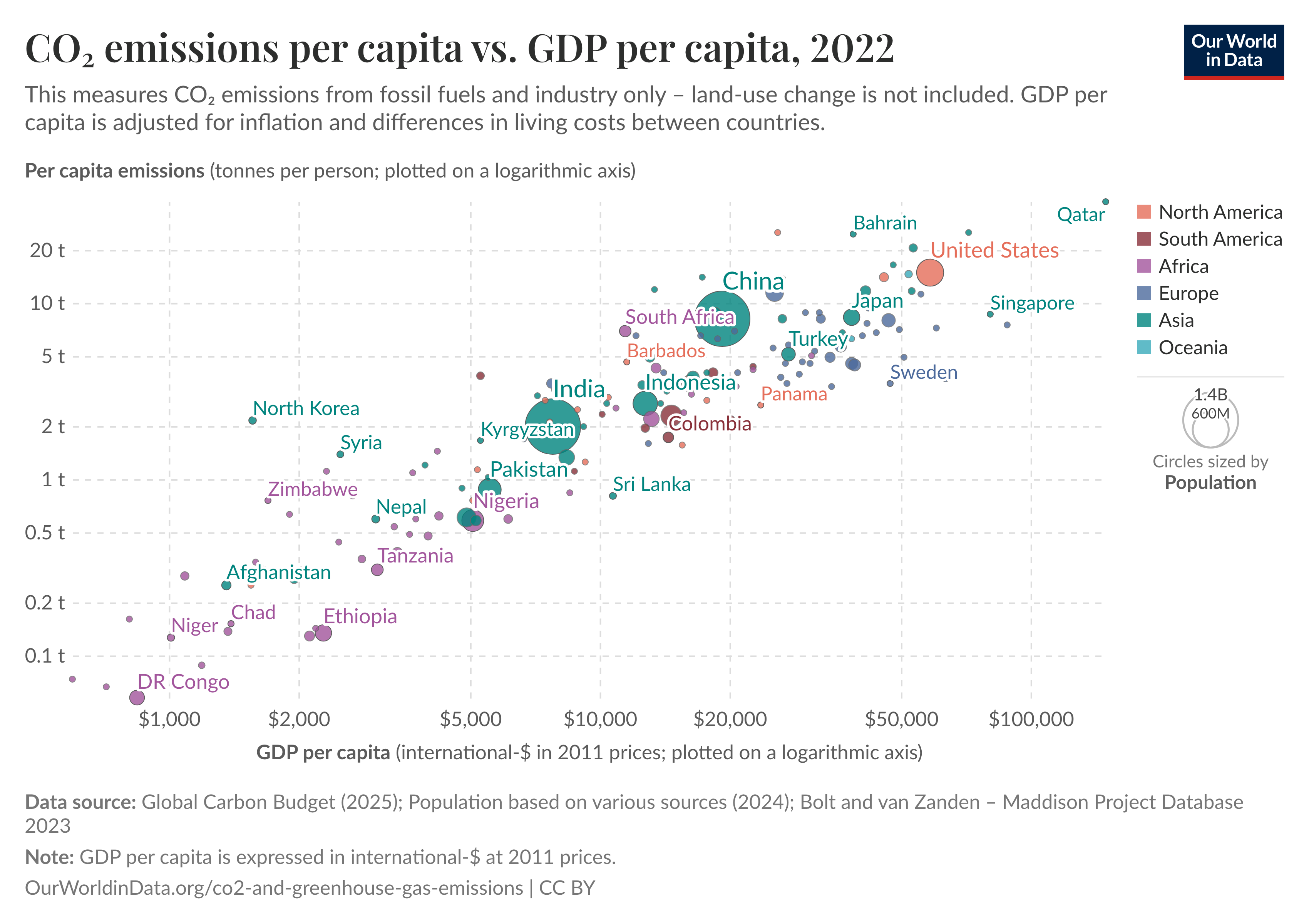}
{adding or removing data points}
{Remove the data point and label for North Korea from the scatter plot.}
{To hide North Korea as a country with relatively high CO2 emissions per capita despite low GDP per capita, thereby weakening the viewer's perception that some lower-income countries can still have comparatively elevated emissions and making the overall relationship appear more uniform.}
{An irregular, \vdbad{squiggly green line was added} to the main plot area, positioned above 'Syria' and below the '2 t' mark on the y-axis.}
{To \vdbad{draw unwarranted attention} to a specific region of the chart, potentially implying a non-existent trend or anomaly among countries with lower GDP and CO2 emissions, thereby misleading the viewer.}
{The data point representing \vagood{North Korea} and its corresponding text label \vagood{were removed} from the scatter plot.}
{To exaggerate the consistency and strength of the positive correlation between national wealth and carbon emissions by hiding a prominent \vagood{low-income, high-emission outlier}.}
{VizAnchor 1.00/1.00; VizDefender 0.00/0.00}
\intentcase
{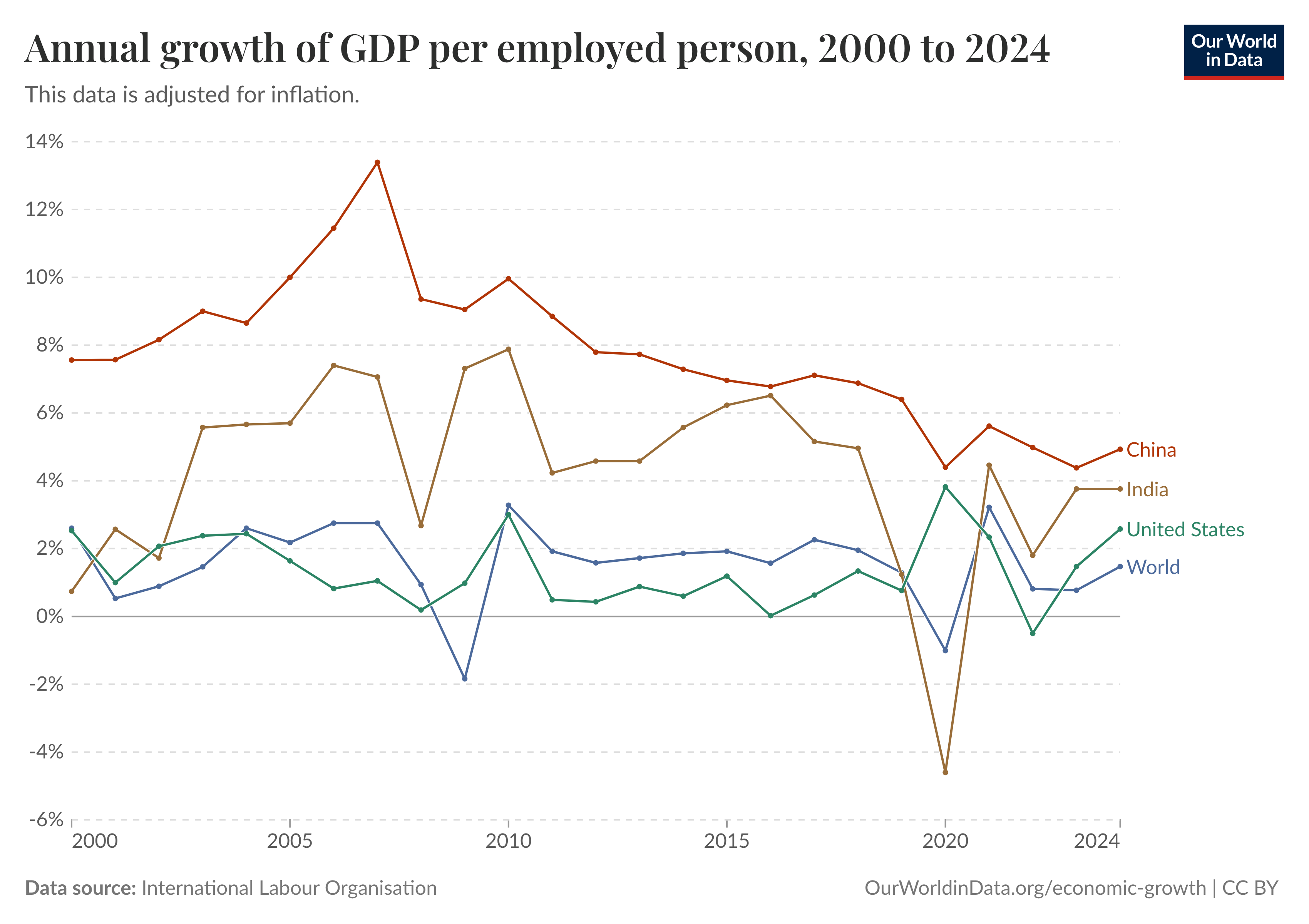}
{modifying coordinate values}
{Change the negative y-axis tick labels from -2\%, -4\%, and -6\% to -5\%, -10\%, and -15\%.}
{To exaggerate the perceived severity of negative GDP-per-employed-person growth, making below-zero downturns appear much larger than the original chart indicates, especially the sharp decline around 2020.}
{The OCR output contains several \vdbad{spurious symbols} that are not present in the visual chart and are clearly misinterpretations of grid lines or tick marks. These are small, irregular shapes, consistent with \vdbad{noise rather than intentional tampering}.}
{The purpose of this region is to highlight \vdbad{OCR noise}, which is not considered intentional tampering. Therefore, there is \vdbad{no malicious intent}.}
{The negative y-axis tick labels \vagood{(-2\%, -4\%, and -6\%)} were modified to \vagood{-5\%, -10\%, and -15\%} respectively, altering the coordinate scale representation on the y-axis while leaving the plotted data lines unchanged.}
{To \vagood{exaggerate the severity of economic contractions}, making India's 2020 GDP growth contraction appear as a massive collapse of over -11\% (instead of -4.6\%) and the World's 2009 recession appear as a drop to nearly -5\% (instead of -1.8\%).}
{VizAnchor 1.00/1.00; VizDefender 0.00/0.00}
\intentcase
{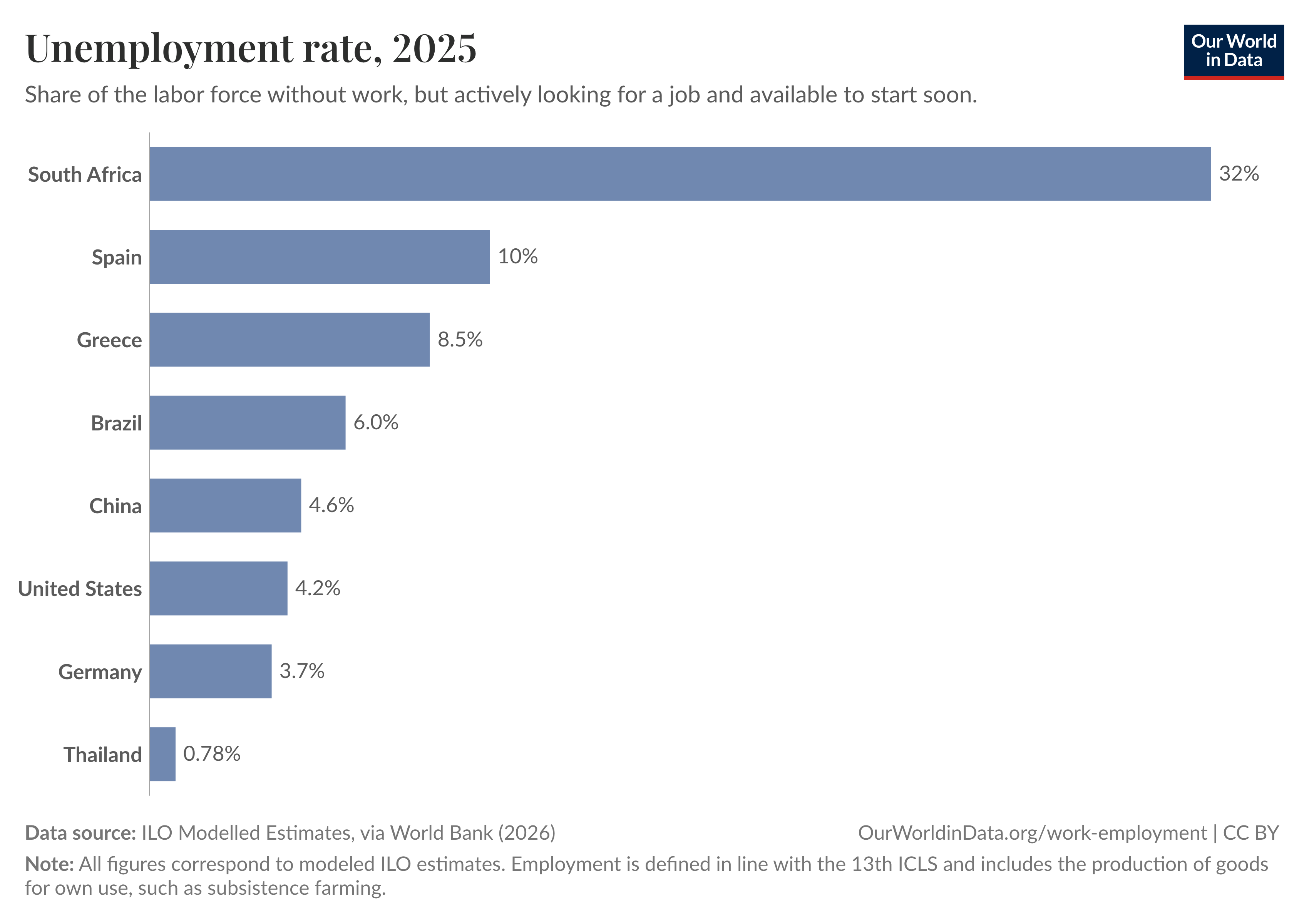}
{deceptive auxiliary annotations}
{Replace the original United States value label "4.2\%" with an oversized bold font.}
{To visually overemphasize the United States' unemployment rate, drawing disproportionate attention to it and making it appear more important or more notable than the other countries shown in the chart.}
{The data label for 'United States' was set to '4.2\%', while its corresponding \vdbad{bar was visually extended} to represent a significantly higher value, creating a mismatch between the numerical data and its visual representation.}
{To deceptively present the unemployment rate for the United States as \vdbad{lower than its actual visual representation}, potentially to portray a more favorable economic situation.}
{The data label \vagood{'4.2\%'} corresponding to the United States bar was \vagood{enlarged significantly} to make it stand out disproportionately compared to the other data labels.}
{To draw \vagood{disproportionate attention} to the United States' 4.2\% unemployment rate, exaggerating its significance or prominence relative to other countries with much higher rates.}
{VizAnchor 0.80/1.00; VizDefender 0.00/0.00}
\intentcase
{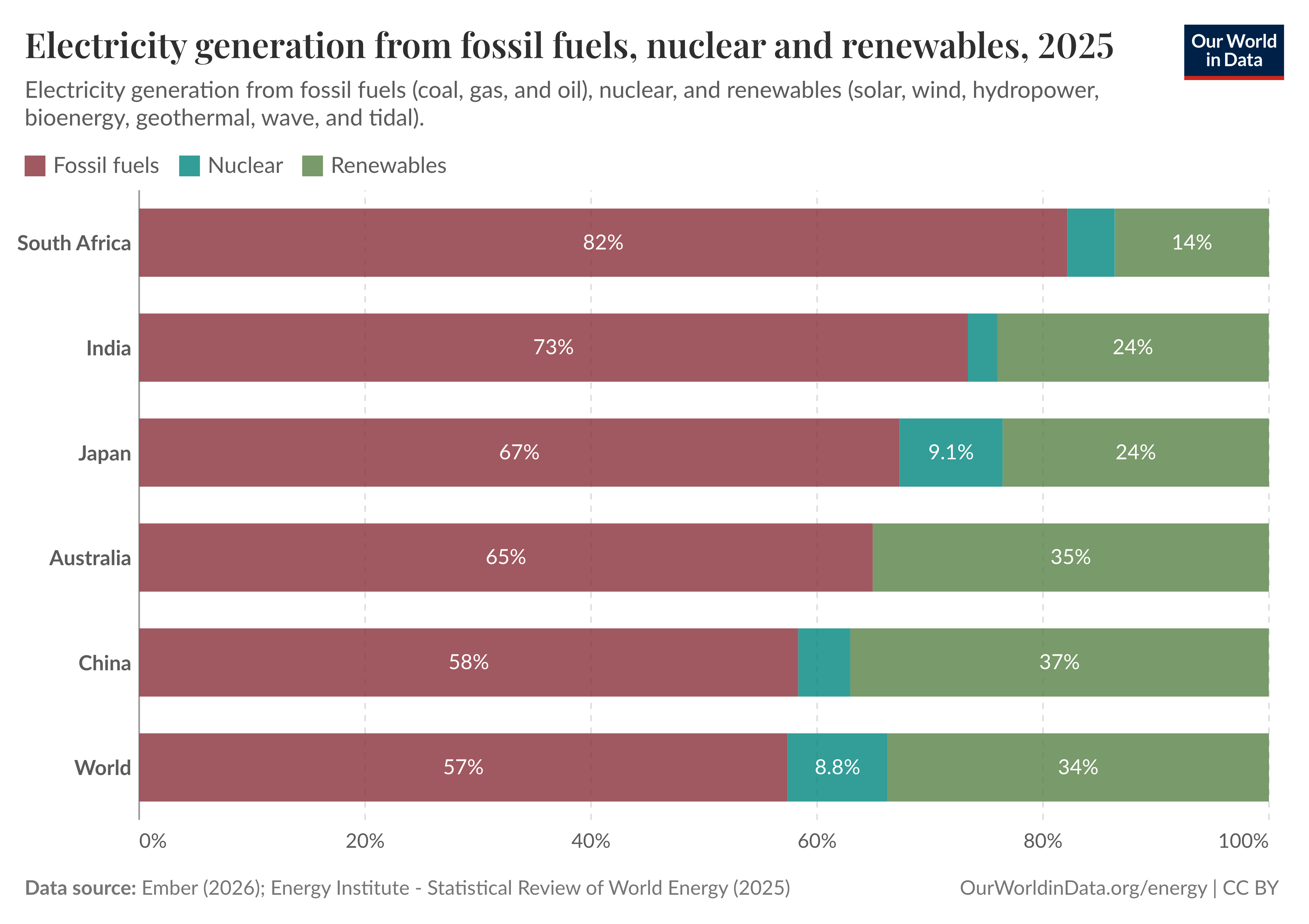}
{modifying the legend}
{Swap the legend labels for Fossil fuels and Renewables.}
{To mislead viewers into interpreting the large fossil-fuel share as renewables and the smaller renewable share as fossil fuels, making the electricity mix appear much cleaner and less dependent on fossil fuels than the original chart shows.}
{The percentage value for \vdbad{Nuclear energy} in Japan's electricity generation was modified to \vdbad{'9.1\%'}.}
{To alter the perceived \vdbad{contribution of nuclear energy} to Japan's electricity generation, potentially to downplay or exaggerate its role in the energy mix.}
{The text labels \vagood{'Fossil fuels' and 'Renewables'} in the legend \vagood{were swapped}, while keeping the color swatches in their original positions. This misrepresents which color corresponds to which category in the stacked bar chart.}
{To falsely depict renewables as the \vagood{dominant source} of electricity generation in 2025 across all shown regions (ranging from 57\% globally to 82\% in South Africa) while portraying fossil fuels as a \vagood{minor contributor}.}
{VizAnchor 1.00/1.00; VizDefender 0.00/0.00}
\intentcase
{}
{hiding labels}
{Fade the numeric value labels for each region and the total value to make them nearly invisible.}
{To obscure the exact number of additional people pushed into poverty in each region and the overall total, making it harder for viewers to accurately compare regional impacts or assess the full scale of the poverty increase.}
{The numerical data values for each region and the total have been \vdbad{altered} from their original figures.}
{To \vdbad{exaggerate the number} of people potentially pushed into poverty by Covid-19, making the impact appear greater than 'half a billion' as stated in the title, by displaying a total of 547.6 million people.}
{The data labels representing the population values at the end of each horizontal bar (e.g., '239.8m', '128.8m', etc.) and the total sum label ('547.6m') inside the 'Total' box have been \vagood{removed/hidden} by blending them into the white background.}
{The likely misleading intent is to obscure the \vagood{exact quantitative scale} of the poverty crisis, preventing viewers from knowing the specific number of people affected in each region and the total global count of 547.6 million.}
{VizAnchor 1.00/1.00; VizDefender 0.00/0.00}
\intentcase
{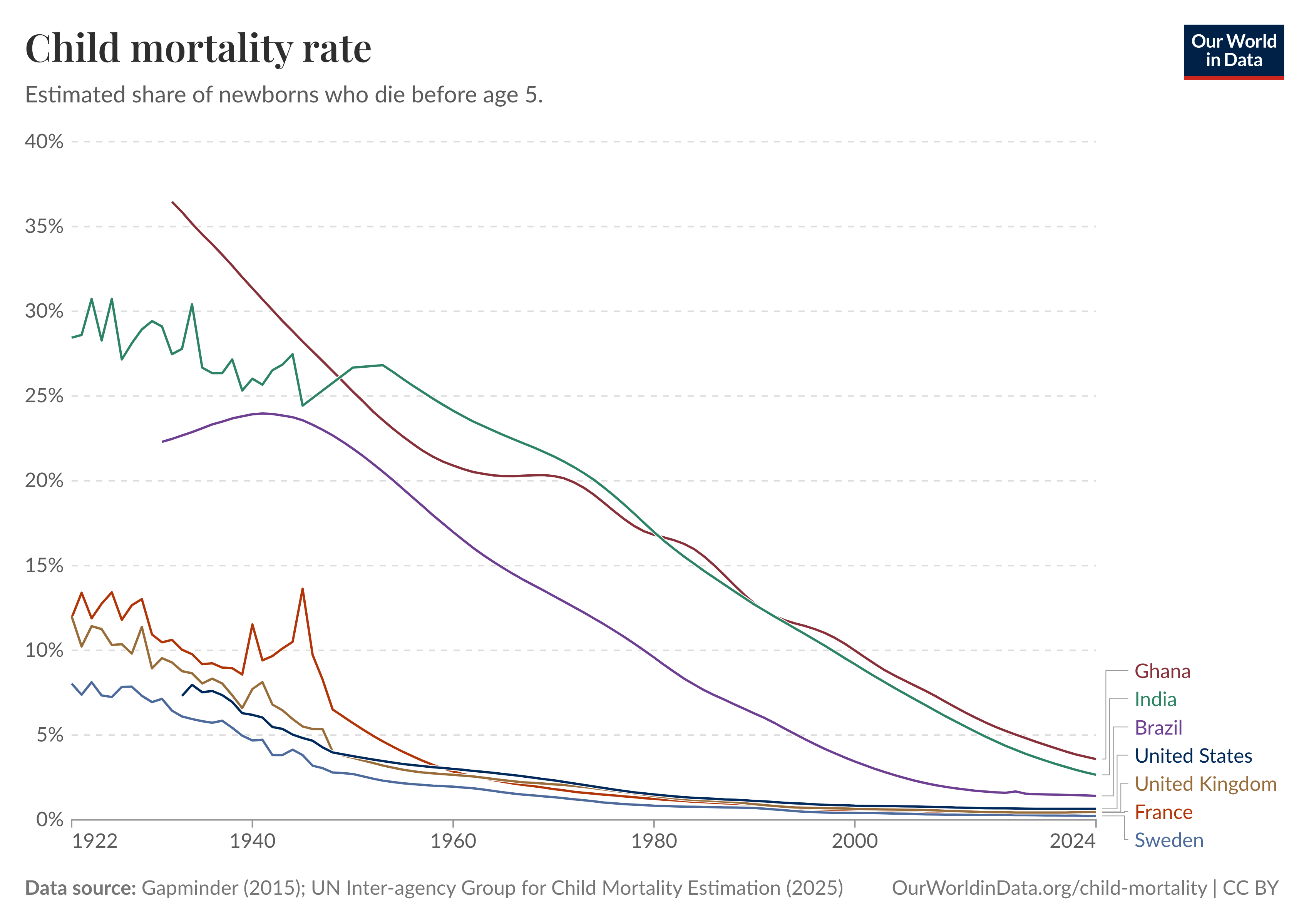}
{adding or removing logos}
{Remove the Our World in Data logo from the top-right corner of the child mortality rate line chart.}
{To obscure the chart's prominent source branding and make the visualization appear less clearly attributable to Our World in Data, reducing viewers' ability to recognize and verify the provenance of the presented child mortality data.}
{The data points for Ghana's child mortality rate between 1922 and 1940 have been \vdbad{artificially increased}, showing a higher and more fluctuating rate than expected for that period, especially compared to the overall downward trend. The data points for India's child mortality rate between 1922 and 1940 have been \vdbad{artificially increased and made more erratic}, showing a higher and more volatile rate than expected for that period. The data points for Brazil's child mortality rate between 1922 and 1940 have been \vdbad{artificially increased}, showing a higher rate than expected for that period, with a peak around 1940.}
{To \vdbad{exaggerate} the initial high child mortality rate in Ghana, potentially to emphasize a more dramatic improvement over time or to highlight a historical disparity. To exaggerate the initial high child mortality rate in India, potentially to emphasize a more dramatic improvement over time or to highlight a historical disparity. To exaggerate the initial high child mortality rate in Brazil, potentially to emphasize a more dramatic improvement over time or to highlight a historical disparity.}
{The \vagood{'Our World in Data' logo} located in the top-right corner of the original chart was \vagood{removed}, leaving a blank white space in the tampered chart.}
{The manipulation is intended to obscure the \vagood{institutional origin and branding} of the chart by removing the 'Our World in Data' logo, thereby hiding the authoritative source of the visualization.}
{VizAnchor 1.00/1.00; VizDefender 0.00/0.00}
\intentcase
{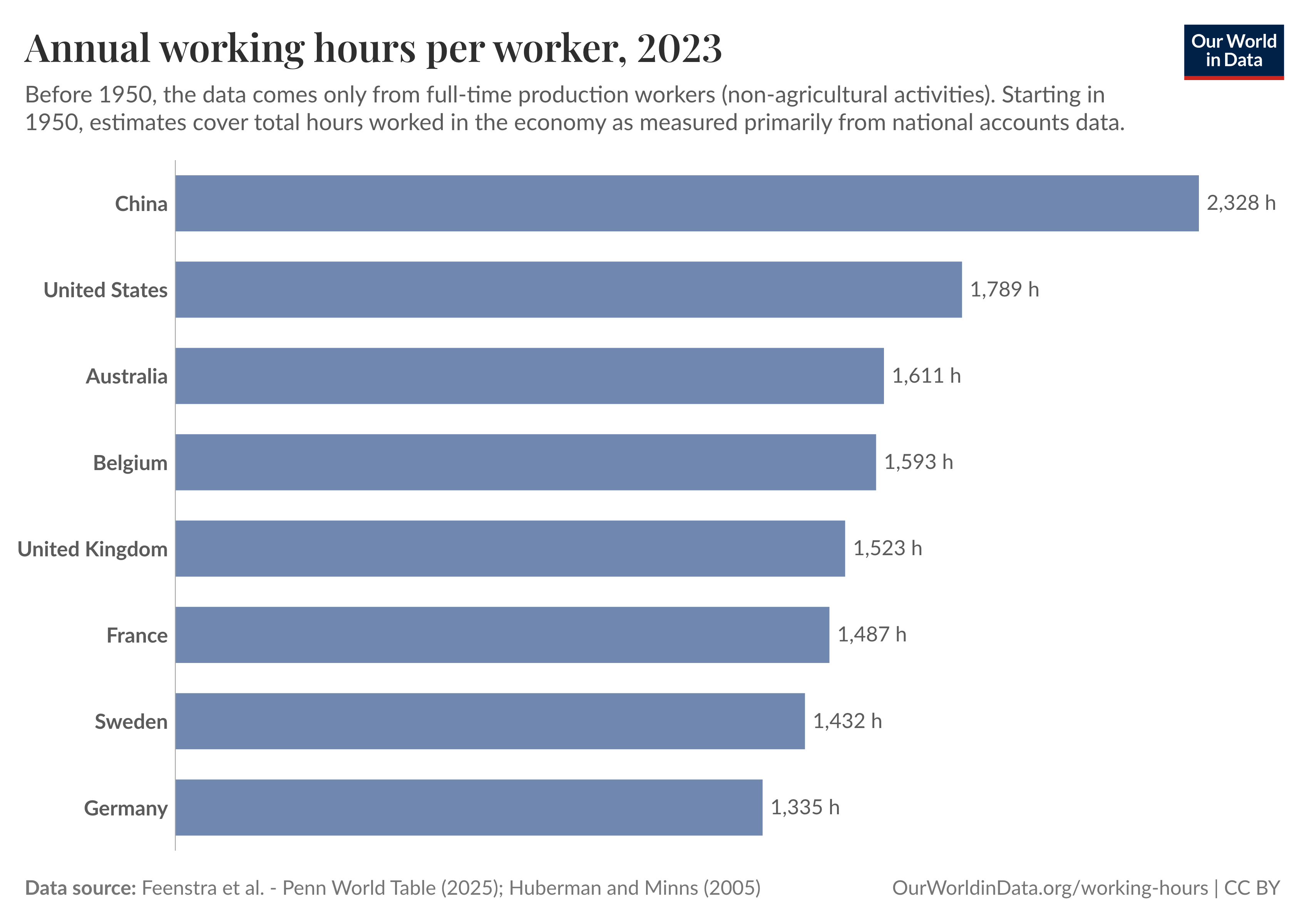}
{data--visual disproportion}
{Shorten the bar for China so its visual length no longer corresponds to the displayed value of 2,328 h.}
{To mislead viewers into underestimating China's annual working hours per worker, making China appear only modestly higher than the United States and closer to other countries than the stated value indicates.}
{The numerical data label for China was \vdbad{increased to '2,328 h'}, while the visual length of its bar was kept shorter, creating a mismatch between the displayed value and its graphical representation.}
{To \vdbad{exaggerate China's annual working hours}, making it appear significantly higher than it actually is, while potentially masking the true visual scale of the data.}
{The horizontal bar representing China was \vagood{shortened} to reduce its visual length, making it appear disproportionately smaller relative to its value. The corresponding data label '2,328 h' was shifted to the left to remain aligned with the end of the shortened bar.}
{To \vagood{downplay the significant gap} in annual working hours between China and other countries, particularly the United States, by making China's working hours appear visually much closer to those of Western nations than the actual data indicates.}
{VizAnchor 1.00/1.00; VizDefender 0.40/0.00}
\intentcase
{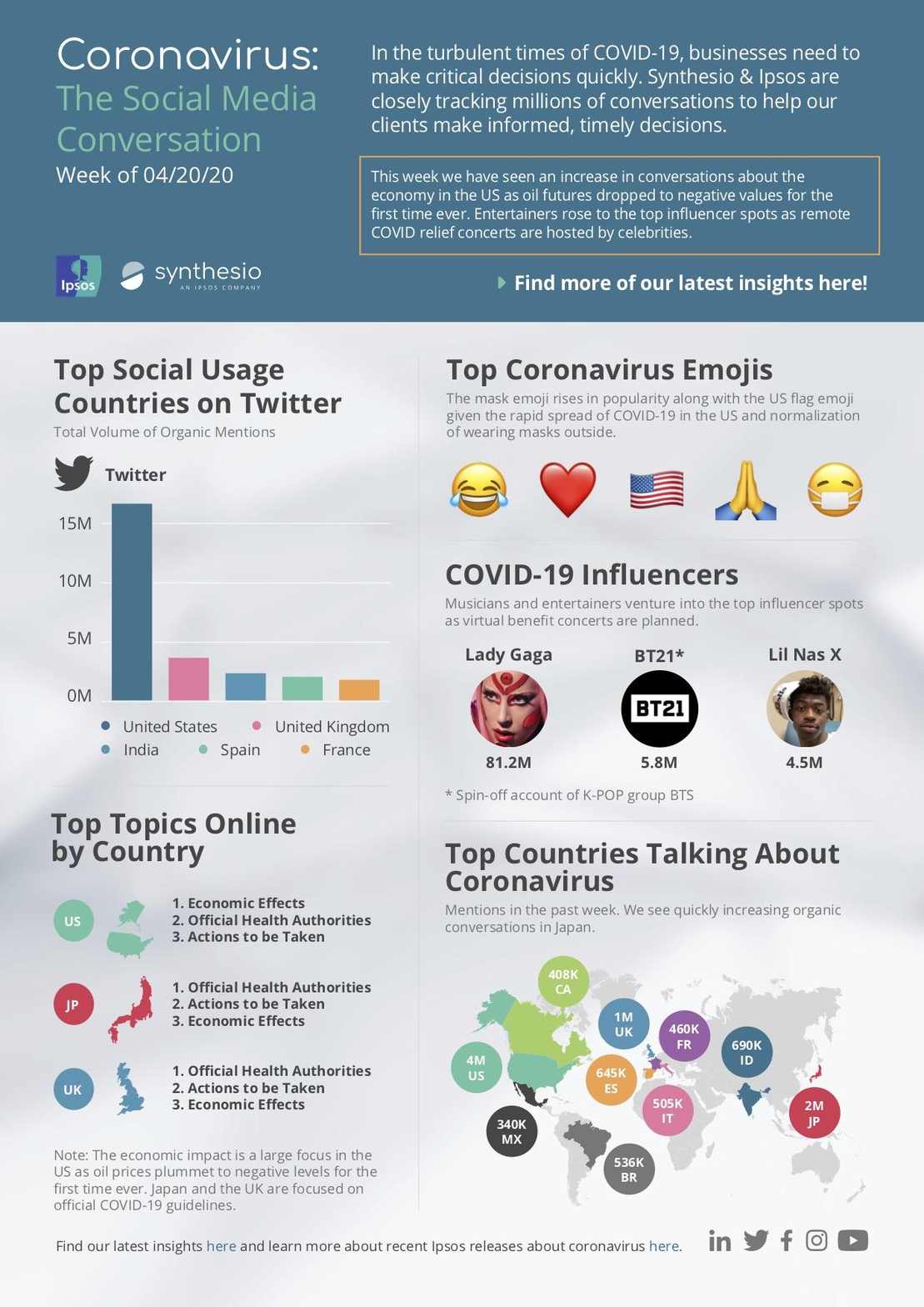}
{modifying the colormap}
{Swap the colors of the United States and United Kingdom bars in the 'Top Social Usage Countries on Twitter' bar chart.}
{To mislead viewers into interpreting the United Kingdom as having the highest Twitter mention volume instead of the United States, thereby distorting the perceived country ranking of COVID-19 social media activity.}
{The \vdbad{'BT21' influencer entry}, including the name 'BT21*', the circular image with 'BT21' text, and the follower count '5.8M'. The green lines clearly highlight the 'BT21' influencer entry within the 'COVID-19 Influencers' section. This area is sufficiently large and encompasses regular shapes (a circular image and text blocks). The edges of the highlighted region are smooth enough to indicate a deliberate selection, consistent with a tampered area rather than noise. The OCR also shows 'BT21*' and 'BT21' which are both highlighted, suggesting a \vdbad{modification related to this specific influencer}.}
{To artificially inflate the perceived diversity or reach of 'influencers' in the COVID-19 conversation by \vdbad{including a non-human, brand-related entity}, potentially to promote the brand or suggest a broader scope of influence than actual musicians and entertainers.}
{The colors of the bars representing \vagood{'United States' and 'United Kingdom'} in the 'Top Social Usage Countries on Twitter' bar chart were \vagood{swapped}. The bar for the United States was changed \vagood{from blue to pink}, and the bar for the United Kingdom was changed \vagood{from pink to blue}, while the legend colors remained unchanged.}
{To falsely convey that the \vagood{United Kingdom had the highest volume} of organic Twitter mentions regarding the coronavirus conversation (exceeding 13 million), while significantly \vagood{downplaying the volume of mentions from the United States} to around 3.5 million.}
{VizAnchor 1.00/1.00; VizDefender 0.00/0.00}
\intentcase
{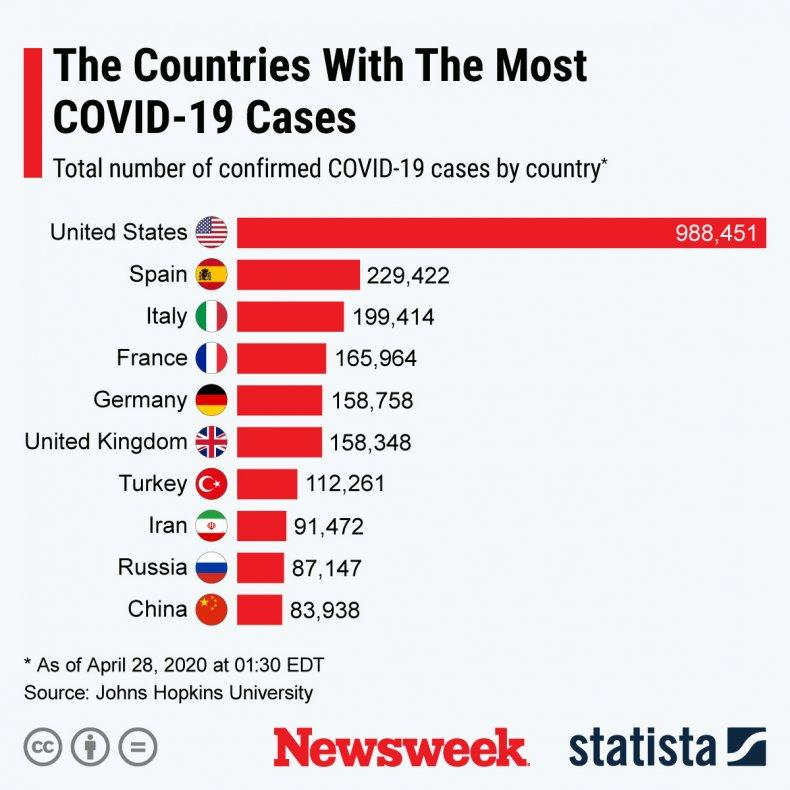}
{mixed tampering}
{Shorten the United States bar so its visual length no longer corresponds to the displayed value of 988,451, and remove the China row from the COVID-19 cases ranking.}
{To mislead viewers into underestimating the scale of confirmed COVID-19 cases in the United States while hiding China from the ranking, making the U.S. appear less dominant and distorting the perceived distribution of cases across countries.}
{The bar representing the United States' COVID-19 cases has been \vdbad{visually extended} to appear significantly longer than its corresponding numerical value of 988,451, creating a visual exaggeration.}
{To \vdbad{exaggerate} the number of COVID-19 cases in the United States by making the bar appear much larger than its actual numerical value, thereby creating a misleading impression of a higher magnitude compared to other countries.}
{The bar representing the United States was \vagood{shortened} to reduce its visual length relative to other countries, while keeping its numerical label. Additionally, the entire row for China was \vagood{completely removed} from the bottom of the chart.}
{To visually \vagood{downplay} the disproportionately high number of COVID-19 cases in the United States relative to other nations, while completely \vagood{concealing China's presence} among the countries with the highest number of confirmed cases as of April 28, 2020.}
{VizAnchor 1.00/1.00; VizDefender 0.00/0.00}

%% file: VizAnchor_ArXiv_Supplement_20260825/complete_main_category_longtables.tex
\begingroup
\centering
\fontsize{6.6}{7.8}\selectfont
\setlength{\tabcolsep}{1.7pt}
\begin{longtable}{@{}p{1.32in}lccccccccc@{}}
\caption{Complete VizDefender and VizAnchor pipeline results by annotated tamper type, pooled over both evaluation datasets. Type Acc. is per-class recall and Type F1 is one-vs-rest class F1. AI scores are normalized to $[0,1]$.}\label{tab:main_tamper_complete}\\
\toprule
\textbf{Category} & \textbf{Method} & \textbf{$n$} & \textbf{Type Acc.} & \textbf{Type F1} & \textbf{Comp. Exact} & \textbf{Comp. F1} & \textbf{Proc. Cos.} & \textbf{Proc. AI} & \textbf{Intent Cos.} & \textbf{Intent AI} \\
\midrule
\endfirsthead
\multicolumn{11}{c}{\tablename\ \thetable{} -- continued} \\
\toprule
\textbf{Category} & \textbf{Method} & \textbf{$n$} & \textbf{Type Acc.} & \textbf{Type F1} & \textbf{Comp. Exact} & \textbf{Comp. F1} & \textbf{Proc. Cos.} & \textbf{Proc. AI} & \textbf{Intent Cos.} & \textbf{Intent AI} \\
\midrule
\endhead
\midrule \multicolumn{11}{r}{Continued on next page} \\
\endfoot
\bottomrule
\endlastfoot
Modifying data-point values & VizDefender & 20 & .600 & .571 & .500 & .362 & .540 & .336 & .722 & .516 \\
 & VizAnchor & 20 & .900 & .878 & .800 & .447 & .674 & .840 & .788 & .830 \\
\addlinespace[1.5pt]
Adding/removing data points & VizDefender & 24 & .708 & .829 & .333 & .622 & .529 & .384 & .641 & .384 \\
 & VizAnchor & 24 & .958 & .958 & .292 & .894 & .734 & .950 & .753 & .750 \\
\addlinespace[1.5pt]
Modifying coordinate values & VizDefender & 20 & .900 & .878 & .400 & .788 & .496 & .250 & .602 & .320 \\
 & VizAnchor & 20 & .950 & .927 & 1.000 & 1.000 & .672 & .940 & .761 & .850 \\
\addlinespace[1.5pt]
Deceptive auxiliary annotations & VizDefender & 20 & .900 & .800 & .500 & .710 & .627 & .550 & .737 & .660 \\
 & VizAnchor & 20 & .900 & .947 & .750 & .857 & .658 & .790 & .780 & .880 \\
\addlinespace[1.5pt]
Modifying legend & VizDefender & 20 & .900 & .947 & .350 & .621 & .547 & .320 & .668 & .320 \\
 & VizAnchor & 20 & .950 & .884 & .550 & .710 & .710 & .980 & .745 & .860 \\
\addlinespace[1.5pt]
Hiding labels & VizDefender & 22 & .909 & .851 & .591 & .753 & .486 & .500 & .707 & .572 \\
 & VizAnchor & 22 & 1.000 & .936 & .773 & .853 & .655 & .964 & .747 & .946 \\
\addlinespace[1.5pt]
Adding/removing logos & VizDefender & 20 & .950 & .950 & .550 & .750 & .532 & .520 & .655 & .690 \\
 & VizAnchor & 20 & 1.000 & .976 & .900 & .974 & .732 & .990 & .723 & .920 \\
\addlinespace[1.5pt]
Data--visual disproportion & VizDefender & 20 & .950 & .792 & .100 & .286 & .711 & .290 & .761 & .350 \\
 & VizAnchor & 20 & .900 & .923 & .450 & .500 & .773 & .810 & .816 & .870 \\
\addlinespace[1.5pt]
Modifying colormap & VizDefender & 20 & .950 & .905 & .100 & .476 & .405 & .120 & .584 & .180 \\
 & VizAnchor & 20 & .650 & .722 & .350 & .702 & .668 & .830 & .691 & .700 \\
\addlinespace[1.5pt]
Mix & VizDefender & 34 & .471 & .582 & .088 & .524 & .529 & .262 & .629 & .450 \\
 & VizAnchor & 34 & .853 & .879 & .529 & .726 & .718 & .924 & .727 & .776 \\
\end{longtable}
\endgroup

\begingroup
\centering
\fontsize{6.6}{7.8}\selectfont
\setlength{\tabcolsep}{1.7pt}
\begin{longtable}{@{}p{1.32in}lccccccccc@{}}
\caption{Complete VizDefender and VizAnchor pipeline results by chart type, pooled over both evaluation datasets. Type F1 is the Macro-F1 within each chart subgroup. AI scores are normalized to $[0,1]$.}\label{tab:main_chart_complete}\\
\toprule
\textbf{Category} & \textbf{Method} & \textbf{$n$} & \textbf{Type Acc.} & \textbf{Type F1} & \textbf{Comp. Exact} & \textbf{Comp. F1} & \textbf{Proc. Cos.} & \textbf{Proc. AI} & \textbf{Intent Cos.} & \textbf{Intent AI} \\
\midrule
\endfirsthead
\multicolumn{11}{c}{\tablename\ \thetable{} -- continued} \\
\toprule
\textbf{Category} & \textbf{Method} & \textbf{$n$} & \textbf{Type Acc.} & \textbf{Type F1} & \textbf{Comp. Exact} & \textbf{Comp. F1} & \textbf{Proc. Cos.} & \textbf{Proc. AI} & \textbf{Intent Cos.} & \textbf{Intent AI} \\
\midrule
\endhead
\midrule \multicolumn{11}{r}{Continued on next page} \\
\endfoot
\bottomrule
\endlastfoot
Bar & VizDefender & 99 & .778 & .778 & .313 & .458 & .578 & .370 & .708 & .472 \\
 & VizAnchor & 99 & .909 & .874 & .515 & .615 & .719 & .912 & .769 & .830 \\
\addlinespace[1.5pt]
Line & VizDefender & 32 & .812 & .758 & .344 & .453 & .511 & .290 & .656 & .394 \\
 & VizAnchor & 32 & .938 & .725 & .812 & .723 & .698 & .862 & .770 & .812 \\
\addlinespace[1.5pt]
Area & VizDefender & 10 & .800 & .829 & .500 & .667 & .579 & .356 & .645 & .444 \\
 & VizAnchor & 10 & .900 & .857 & .800 & .964 & .694 & .840 & .744 & .860 \\
\addlinespace[1.5pt]
Scatter & VizDefender & 16 & .812 & .708 & .312 & .410 & .485 & .212 & .631 & .324 \\
 & VizAnchor & 16 & .875 & .833 & .812 & .884 & .682 & .888 & .714 & .862 \\
\addlinespace[1.5pt]
Map & VizDefender & 29 & .862 & .860 & .276 & .408 & .486 & .352 & .659 & .442 \\
 & VizAnchor & 29 & .897 & .906 & .655 & .653 & .674 & .952 & .731 & .842 \\
\addlinespace[1.5pt]
Pie/Donut & VizDefender & 8 & .625 & .575 & .375 & .607 & .566 & .500 & .562 & .550 \\
 & VizAnchor & 8 & .875 & .900 & .250 & .480 & .680 & .900 & .687 & .800 \\
\addlinespace[1.5pt]
Other/Mixed & VizDefender & 26 & .846 & .838 & .423 & .452 & .491 & .384 & .602 & .440 \\
 & VizAnchor & 26 & .885 & .898 & .731 & .704 & .687 & .916 & .728 & .846 \\
\end{longtable}
\endgroup